\documentclass[nonacm,sigconf,screen,balance=false]{acmart}
\usepackage[utf8]{inputenc}
\usepackage[T1]{fontenc}
\usepackage{microtype}
\usepackage{subcaption}

\setcitestyle{nosort}

\definecolor[named]{xBlue}{HTML}{18647E}
\definecolor[named]{xOrange}{HTML}{FF9B00}
\definecolor[named]{xGray}{HTML}{808080}
\definecolor[named]{xRed}{HTML}{A30B37}

\definecolor[named]{cPink}{HTML}{F26DF9}

\def\Snospace~{\S{}}

\usepackage{epigraph}
\usepackage[hang,flushmargin]{footmisc}

\usepackage{float}
\usepackage{dblfloatfix}

\renewcommand{\bibliofont}

\newcommand{\Voltron}{$\mathcal{V}$oltron}  
\newcommand{\VCond}{\textbf{$\mathcal{V}$ -- Cond}}
\newcommand{\VDual}{\textbf{$\mathcal{V}$ -- Dual}}
\newcommand{\VGen}{\textbf{$\mathcal{V}$ -- Gen}}

\newcommand{\unfair}[1]{{\color{xGray}\textit{#1}}}

\begin{document}

\title[\Voltron{}: Language-Driven Representations for Robotics]{
    Language-Driven Representation Learning for Robotics
}

\author{Siddharth Karamcheti}
\affiliation{
    \institution{Stanford University}
}
\email{skaramcheti@cs.stanford.edu}

\author{Suraj Nair}
\affiliation{
    \institution{Stanford University}
}
\email{surajn@cs.stanford.edu}

\author{Annie Chen}
\affiliation{
    \institution{Stanford University}
}
\email{asc8@cs.stanford.edu}

\author{
    \medskip
    \hspace*{-14.75pc}
    \makebox[\textwidth]{
        \makebox[1pc]{}
        \makebox[9.2978pc]{Thomas Kollar}
        \makebox[1.1pc]{}
        \makebox[9.2978pc]{Chelsea Finn}
        \makebox[1.1pc]{}
        \makebox[9.2978pc]{Dorsa Sadigh}
        \makebox[1.1pc]{}
        \makebox[9.2978pc]{Percy Liang}
        \makebox[1pc]{}
    }
}
\affiliation{
    \hspace*{-14.75pc}
    \makebox[\textwidth]{
        \makebox[1pc]{}
        \makebox[9.2978pc]{Toyota Research Institute}
        \makebox[1.1pc]{}
        \makebox[9.2978pc]{Stanford University}
        \makebox[1.1pc]{}
        \makebox[9.2978pc]{Stanford University}
        \makebox[1.1pc]{}
        \makebox[9.2978pc]{Stanford University}
        \makebox[1pc]{}
    }
    \medskip
}

\renewcommand{\shortauthors}{Karamcheti et. al.}

\begin{abstract}
Recent work in visual representation learning for robotics demonstrates the viability of learning from large video datasets of humans performing everyday tasks. Leveraging methods such as masked autoencoding and contrastive learning, these representations exhibit strong transfer to policy learning for visuomotor control. But, robot learning encompasses a diverse set of problems beyond control including grasp affordance prediction, language-conditioned imitation learning, and intent scoring for human-robot collaboration, amongst others. First, we demonstrate that existing representations yield inconsistent results across these tasks: masked autoencoding approaches pick up on low-level spatial features at the cost of high-level semantics, while contrastive learning approaches capture the opposite. We then introduce \Voltron{}, a framework for language-driven representation learning from human videos and associated captions. \Voltron{} trades off language-conditioned visual reconstruction to learn low-level visual patterns, and visually-grounded language generation to encode high-level semantics. We also construct a new evaluation suite spanning five distinct robot learning problems -- a unified platform for holistically evaluating visual representations for robotics. Through comprehensive, controlled experiments across all five problems, we find that \Voltron{}'s language-driven representations outperform the prior state-of-the-art, especially on targeted problems requiring higher-level features.\footnote{
Project Page: \url{https://sites.google.com/view/voltron-robotics}\\
Model Artifacts \& Pretraining Code: \url{https://github.com/siddk/voltron-robotics}\\
Evaluation Suite: \url{https://github.com/siddk/voltron-evaluation}
}

\end{abstract}

\maketitle

\section{Introduction}
\label{sec:introduction}

\begin{figure*}[t]
    \centering
    \includegraphics[width=\linewidth]{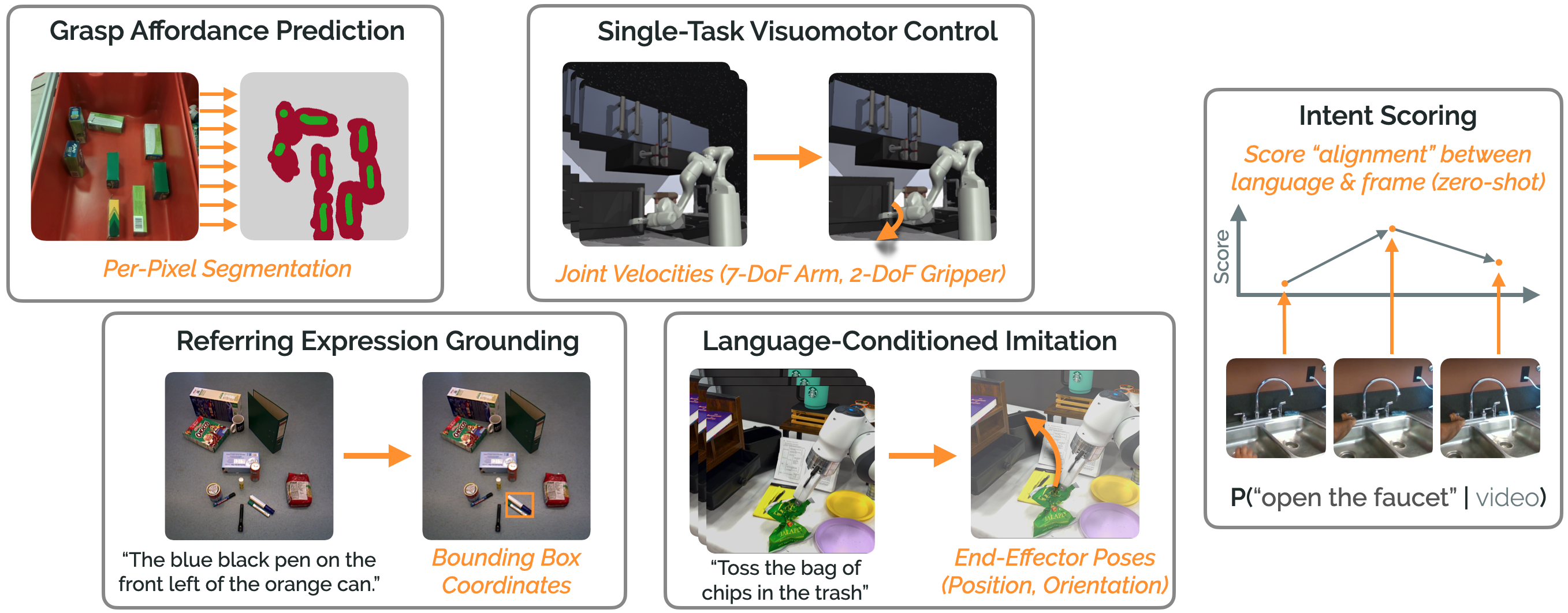}
    \vspace*{-5mm}
    \caption{\textbf{\Voltron{} Evaluation Suite.} We introduce a suite of evaluation problems spanning \textit{five applications within robotics}, including grasp affordance prediction, referring expression grounding, single-task visuomotor control (in simulation), language-conditioned imitation learning (on a real robot), and intent scoring.}
    \label{fig:front-fig}
    \vspace*{-3mm}
\end{figure*}

\epigraph{
    \textit{Good words are worth much, and cost little.}
}{
    \textsc{--- George Herbert}
}
\begin{center}\rule{0.7\linewidth}{0.3pt}\end{center}
\vspace*{1mm}


Realizing a future of ubiquitous, broadly capable robots is predicated on systems capable of generalizable perception and interaction \citep{weiss1987dynamic, chaumette2006visualservo, levine2016end}. Towards this goal, recent work in robotics present approaches for learning visual representations to bootstrap learning for visuomotor control \citep{parisi2022unsurprising, nair2022r3m, radosavovic2022mvp}. Critically, these approaches show  that we can learn such representations from \textit{real-world videos of human behavior} -- specifically, egocentric video datasets such as Something-Something-v2 and Ego4D \citep{goyal2017sthsth, grauman2022ego4d} -- instead of solely relying on in-domain robotics data that is scarce and expensive. While prior work has developed and evaluated representations for visuomotor control, robot learning is an expansive discipline, spanning \textit{a diverse spectrum of problems}: predicting grasp proposals from visual input \citep{saxena2008visualgrasping, mahler2017dexnet2}, language-conditioned imitation learning \citep{tellex2011understanding} and belief/intent tracking for human-robot interaction \citep{hauser2012recognition, javdani2018shared}, amongst others. Broadening our focus to problems beyond learning for control enables us to develop flexible, generalizable representations that capture \textit{both} low-level spatial reasoning and high-level semantic understanding -- a flexibility that is a key prerequisite to realizing a foundation model for robotics \citep{bommasani2021opportunities}. Thus, we ask: \textit{how can we learn visual representations that generalize across the diverse spectrum of problems in robot learning?}

Recent approaches for learning visual representations for robotics use pretraining objectives that reflect different inductive biases for what the learned representations should capture. Masked Visual Pretraining \citep[MVP;][]{radosavovic2022mvp} proposes using masked autoencoding \citep{he2022mae} to prioritize visual reconstruction from heavily masked video frames, encoding representations that facilitate per-pixel reconstruction. Separately, Reusable Representations for Robotic Manipulation \citep[R3M;][]{nair2022r3m} eschews pixel reconstruction for two contrastive learning objectives: time contrastive learning \citep{sermanet2018tcn} and video-language alignment. These approaches show strong performance on imitation learning in simulated and real-world settings, with sizeable improvements over strong alternatives such as ResNet or CLIP features \citep{he2016resnet, radford2021clip}; however, they have not been evaluated beyond these settings. As a first contribution, we evaluate these representations on problems beyond control and identify \textit{inconsistent evaluation performance}, with huge penalties depending on the approach and specific application. MVP performs well on problems such as grasp affordance prediction, but struggles with higher-level problems such as language-conditioned imitation. R3M instead excels at the higher-level problems, but degrades completely on problems such as grasp affordance prediction.

Motivated by this, we present \textbf{\Voltron{}}, a framework for language-driven visual representation learning for robotics that learns representations that capture both low-level and high-level features, empirically outperforming prior approaches over \textit{all} applications. \Voltron{} models take videos and associated language captions as input to a masked autoencoding pipeline, reconstructing one (or more) frames from a masked context. \textit{The novelty of our framework is in how we use language supervision}. Depending on a tunable probability $\alpha$, we either condition on $(\alpha = 0)$, or generate ($\alpha > 0)$ the associated caption. Explicitly \textit{conditioning} on words in different contexts allows for low-level pattern recognition at the local, spatial level, while \textit{generating} language from our learned visual encoding allow us to infer higher-level features around affordances and intents. Furthermore, guided by the hypothesis that language is especially useful in describing \textit{change}, we study \textit{dual-frame} contexts consisting of the initial and current observation in multi-timestep tasks. Altogether, we examine \textit{three different \Voltron{} variants}: \VCond{} (\textit{Language Conditioning}: single frame, $\alpha = 0$), \VDual{} (\textit{Adding Context}: dual-frame conditioning, $\alpha = 0$), and \VGen{} (\textit{Adding Language Generation}: dual-frame, $\alpha = 0.5$ -- we find that $\alpha = 1$ with no language-conditioning at all hurts performance). 

To evaluate \Voltron{} and other visual representation learning approaches, we assemble a new evaluation suite (depicted in \autoref{fig:front-fig}) spanning five problem domains within robotics: 1) dense segmentation for grasp affordance prediction \citep{zeng2017arcgrasping}, 2) object detection from referring expressions (e.g., ``the blue coffee mug to the left of the plate'') in cluttered scenes \citep{wang2021ocidref}, 3) imitation learning for visuomotor control (in simulation) \citep{nair2022r3m}, 4) learning multi-task language-conditioned policies for real-world manipulation \citep{stepputtis2020lcil} (on a real-world Franka Emika fixed-arm manipulator), and 5) zero-shot intent scoring \citep{javdani2018shared, chen2021generalizable}. We choose these tasks for their broad coverage; tasks such as grasp affordance prediction and referring expression grounding require reasoning over low-level spatial features, while language-conditioned imitation and intent scoring require a deeper understanding of semantics. 

Through experiments controlling for pretraining data and model capacity, we show that the simplest \Voltron{} representations (from \VCond{}) strictly outperform both MVP and R3M representations across \textit{all} evaluation domains. Furthermore, by adapting our models to learn from multiple frame contexts and that favor generation (e.g., with \VDual{} and \VGen{}), we show that we can further boost performance on evaluations requiring higher-level features such as with language-conditioned policy learning (on a real robot) and intent scoring. Though language-conditioning offers universal performance gains, there are tradeoffs \textit{between \Voltron{} models}; adding language generation hurts performance on some control tasks, even though its necessary for strong performance on intent scoring. Furthermore, \Voltron{} with single-frame language conditioning performs well on non-episodic tasks (e.g., grasping), but underperforms multi-frame models on control tasks. There is not yet a silver bullet -- a single representation strong on all tasks -- but the ability to balance tradeoffs between encoding low and high-level features offers a net win over restrictions of past work.

\smallskip

\noindent \textbf{Contributions.} \textsc{1)} We present \textbf{\Voltron{}}, a framework for language-driven visual representation learning. Through controlled experiments and comprehensive ablations we demonstrate that \Voltron{}'s representations strictly outperform the prior art across \textsc{2)} a new evaluation suite composed of five distinct problem domains within robotics. Finally, \textsc{3)} we analyze the tradeoffs between different \Voltron{} models that balance different types of feature learning, outlining several directions for future work. We release all models, the evaluation suite, code (pretraining and adaptation), and preprocessed data (\url{https://sites.google.com/view/voltron-robotics}).

\smallskip

\noindent \textbf{Limitations.} We do not have access to the compute resources to train models of the same scale and data used in prior work \citep{radosavovic2022mvp, nair2022r3m}. Instead, we carefully reproduce MVP and R3M -- the current state-of-the-art approaches -- by pretraining on the Something-Something-v2 dataset \citep{goyal2017sthsth}, further controlling for batch ordering, model capacity, and other sources of randomness (full details are in \autoref{sec:implementation-reproducibility}). However, for full context we also include results from the official release artifacts from both these works, as well as other methods such as CLIP \citep{radford2021clip}, though we note these results \unfair{in gray} or with dashed lines as to indicate they are not directly comparable.

\section{Related Work}
\label{sec:related-work}

\begin{figure*}[t]
    \centering
    \includegraphics[width=\linewidth]{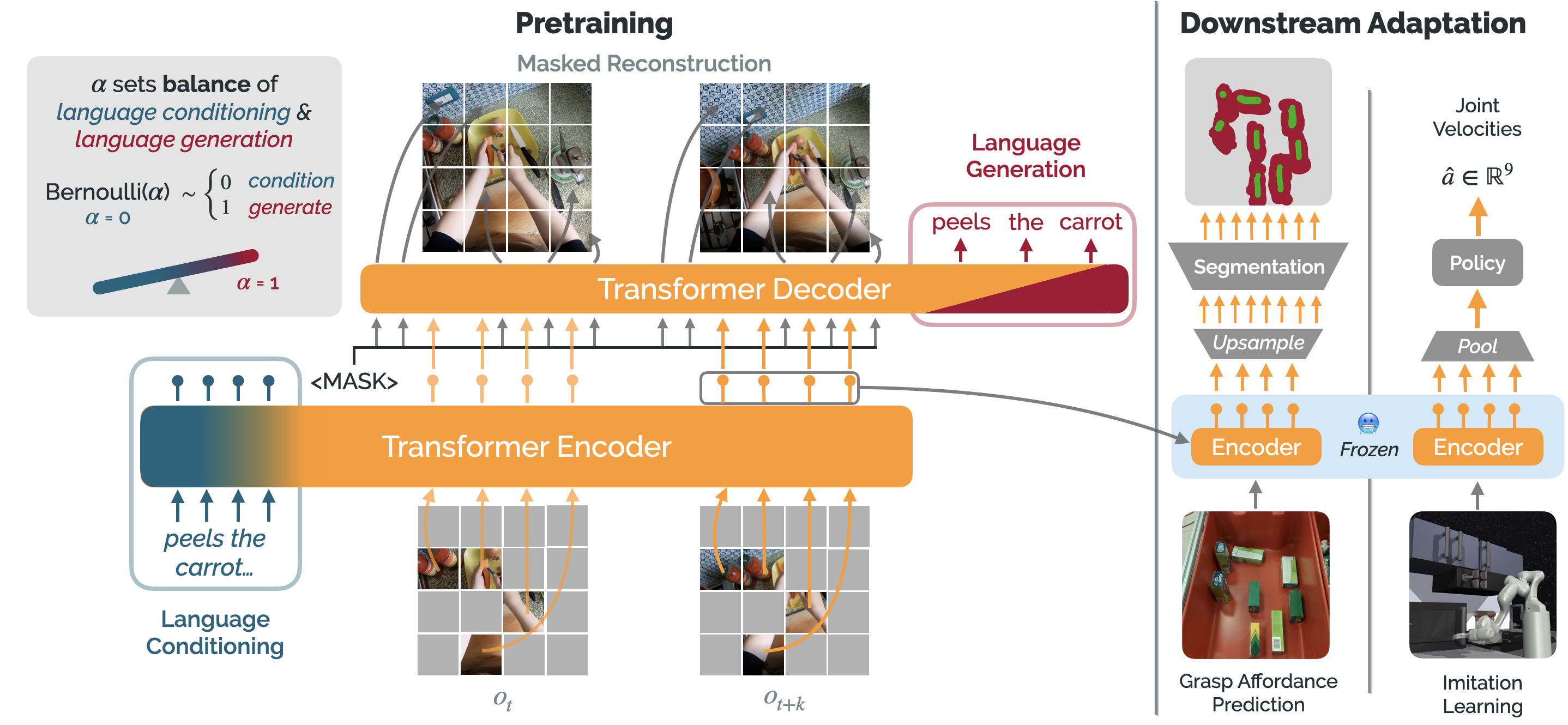}
    \vspace*{-5mm}
    \caption{\textbf{The \Voltron{} Framework}. Central to our approach is \textit{language-driven learning} on top of a masked autoencoding backbone. We incorporate language in two ways, following \autoref{subsec:voltron-balancing}: 1) as a \textit{conditioning variable} fed to a multimodal encoder that also encodes one or more video frames, or 2) as a \textit{generation target} for the language generator \textbf{[Left]}. During downstream evaluation, we use the (frozen) outputs from the encoder, adapting evaluation-specific ``heads'' on top \textbf{[Right]}.}
    \label{fig:voltron-framework}
\end{figure*}


\Voltron{} is situated within a rich body of work in visual representation learning for robotics and multimodal pretraining.

\smallskip

\noindent \textbf{Visual Representation Learning for Robotics.} An emerging body of work in robot learning studies learning visual state representations for control. A wealth of prior approaches learn representations from \textit{in-domain} data taken directly from the target environment (and corresponding task); these techniques range from using data augmentation \citep{laskin2020reinforcement, srinivas2020curl, kostrikov2021image, pari2022surprising} to modeling forward dynamics \citep{gelada2019deepmdp, hafner2020dream2control} to using task-specific information \citep{jonschkowski2015statereps, zhang2021invariant}. Unlike these approaches, we move beyond task-specific data, instead leveraging large, accessible datasets such as videos of humans performing everyday tasks. Work in this paradigm has exploded in recent years. A number of approaches find that existing representations such as features from models trained on ImageNet \citep{deng2009imagenet}, or features from CLIP \citep{radford2021clip} enable more efficient learning \citep{shah2021rrl, khandelwal2021simple}. More recently, multiple approaches have shown increased dividends in applying such representations to visuomotor control, for example by combining features at different layers of pretrained ResNets \citep{parisi2022unsurprising} or by pretraining such representations on human videos, conjecturing that such data captures features useful for robotic manipulation \citep{nair2022r3m, xiao2022mvp, radosavovic2022mvp, ma2022vip}. However, missing from these approaches is a notion of semantics; works such as MVP \cite{xiao2022mvp, radosavovic2022mvp} purely learn to perform masked reconstruction from a single image, and even works that leverage \textit{some} temporal and linguistic signals do so in a limited way \citep{nair2022r3m, ma2022vip}. Instead, our work is motivated by the hypothesis that language understanding -- both via conditioning \textit{and} generation -- is an essential component of learning generalizable visual representations. It is not enough that a representation summarizes an observation; instead, for generalization to new contexts and behaviors, it must capture how observations (and \textit{changes} thereof) relate to higher-level semantic abstractions. 

\Voltron{} aims to do this with its language-driven representation learning objective: by jointly modeling sequences of frames \textit{and} language, we enable a range of capabilities, from producing representations of single images in isolation, to providing the capability to \textit{generate} language grounded in visual contexts. We demonstrate the benefits of language-driven learning in our evaluation (see \autoref{sec:evaluation-suite-results}): in head-to-head comparisons, \Voltron{} models strictly outperform prior approaches across \textit{all} evaluation domains. 

\smallskip

\noindent \textbf{Learning Multimodal Foundation Models.} Our work draws further inspiration from a wave of progress in multimodal foundation models such as CLIP, Multimodal Masked Autoencoders (M3AE), Flamingo, CoCa, and Gato, amongst many others \citep{radford2021clip, geng2022m3ae, alayrac2022flamingo, yu2022coca, reed2022gato, lu2023unifiedio, aghajanyan2022cm3}. These approaches highlight the myriad benefits of multimodal pretraining: language supervision works to enrich visual representations (even \textit{in the absence of language downstream}), while visual supervision similarly enriches language representations \citep{lu2019vilbert, singh2022flava}. Of the many capabilities afforded by these models, many have applications in embodied AI and robotics. CLIP representations have shown to be effective in applications to various robotics tasks \citep{shridhar2021clipport, khandelwal2021simple, cui2022zeroshot}, while multimodal transformer models have proven effective initializations for training control policies \citep{reid2022wikipedia, liu2022instructrl}. These approaches are similar to \Voltron{} in their joint use of visual and language inputs; where \Voltron{} differs, however, is in our novel representation learning objective that balances language conditioning and generation, enabling learning representations that transfer to a wide range of applications within robotics.

\section{\Voltron{} -- Language-Driven Learning}
\label{sec:voltron}


We assume access to a dataset of videos paired with natural language annotations; in each video-language pair $(v, c)$, language can take the form of a caption (e.g., ``peels the carrot'' in \autoref{fig:voltron-framework}), narration, or even coarse textual label of a behavior. We assume each video $v \in \mathbb{R}^{T \times H \times W \times C}$ consists of a sequence of frames $v = [o_{1}, \dots, o_{T}]$, where each frame $o_{i} \in \mathbb{R}^{H \times W \times C}$ is RGB-encoded. We tokenize and one-hot encode each utterance into a vocabulary $V$ of cardinality $|V|$, padding to a max length $L$ such that $c \in \mathbb{R}^{L \times |V|}$. We define a \texttt{<NULL>} token (separate from the \texttt{<PAD>} token) as a placeholder for an empty language context. Furthermore, following the MAE work, we define a visual masking function $\texttt{Mask}(v, \gamma) \rightarrow (v_\text{visible} \in \mathbb{R}^{(1 - \gamma)(T \times H \times W \times C)}, v_\text{masked} \in \mathbb{R}^{\gamma(T \times H \times W \times C)})$ that partitions the regions of a video into a set of visible and masked-out regions subject to a fixed masking ratio $\gamma$. We sample a mask once, and apply it uniformly across \textit{all} frames in the video to prevent leakage \citep{tong2022videomae}; if the masks were sampled independently, a masked region in one frame could be visible in another, allowing the encoder to ``cheat'' by looking ahead.

\subsection{\Voltron{} -- Core Components}

A \Voltron{} model comprises 1) a \textit{multimodal encoder} that takes in a visual context and (optional) language utterance producing a dense representation, 2) a \textit{visual reconstructor} that attempts to reconstruct the masked-out visual context from the encoder's representation of what is visible, and 3) a \textit{language generator} that predicts the language annotation for the video given the encoded visual context. The visual reconstructor and language generator crucially act to shape the representations by first erasing portions of a $(v, c)$ pair, then attempting to reconstruct the missing parts; we show in our experiments (see \autoref{sec:evaluation-suite-results}) that this bottleneck helps focus on more low-level features when we favor reconstruction over generation, and more high-level, semantic features when we favor generation over reconstruction. We step through each component below.

\medskip

\noindent \textbf{Multimodal Encoder:} $\text{E}_\theta(\tilde{v}, u) \rightarrow h \in \mathbb{R}^{S \times d}$ 

\smallskip
\noindent The multimodal encoder (\autoref{fig:voltron-framework}; lower half in \textcolor{xBlue}{blue} and \textcolor{xOrange}{orange}) is the core of a \Voltron{} model. It takes as input $(\tilde{v}, u)$ where $\tilde{v} \in \{v_\text{visible}, v\}$ denotes either the \textit{masked} or \textit{unmasked} (full) visual context respectively, and $u$ represents a (possibly \texttt{<NULL>}) utterance to condition on. As output, the encoder produces a dense representation $h \in \mathbb{R}^{S \times d}$ where $S$ denotes the number of encoded regions, and $d$ is a hyperparameter denoting the dimensionality of the representation. Keeping with the original MAE work, we divide each image $o_i \in \mathbb{R}^{H \times W \times C}$ into a set of non-overlapping regions $R$, where each region is a $p \times p$ patch; this results in $|R| = HW / p^2$ regions. Given a $k$-frame context, $S = (1 - \gamma) k |R|$.

\medskip

\noindent \textbf{Visual Reconstructor}: $\text{R}_\theta(h) \rightarrow \hat{v}_\text{masked} \in \mathbb{R}^{\gamma(k \times H \times W \times C)}$

\smallskip
\noindent The visual reconstructor (\autoref{fig:voltron-framework}; upper half in \textcolor{xOrange}{orange}) takes as input the encoded representation of the \textit{visible} visual context $h = \text{E}_\theta(v_\text{visible}, c)$. It attempts to reconstruct the missing visual regions $v_\text{masked}$, conditioned on language context $c$, producing a prediction $\hat{v}_\text{masked}$. Following prior work, the elements of $\hat{v}_\text{masked}$ are the normalized pixel targets from the original image. We use mean-squared error as the reconstruction loss $\mathcal{L}_\text{reconstruct}(\theta)$.

\medskip

\noindent \textbf{Language Generator}: $\text{G}_\theta(h) \rightarrow \hat{c} \in \mathbb{R}^{L \times C}$

\smallskip
\noindent The language generator (\autoref{fig:voltron-framework}; upper half in \textcolor{xRed}{red}) takes the encoded representation of the \textit{visible} context and the \texttt{<NULL>} language token, $h = \text{E}_\theta(v_\text{visible}, \texttt{<NULL>})$. It generates the language annotation, producing $\hat{c} \in \mathbb{R}^{L \times |V|}$, with each of the $L$ elements corresponding to a probability distribution over the vocabulary. We use the negative log-likelihood (cross-entropy) of the annotation $c$ under the generator as our loss $\mathcal{L}_\text{generate}$.

The language generator crucially takes the \texttt{<NULL>} token as input instead of the annotation $c$; inputting the same $c$ that the generator is trying to output can lead to trivial collapse where the encoder learns to memorize the tokens to aid the generator. As a result, for each example during training we need to \textit{either} condition \textit{or} generate language; this further motivates the parameter $\alpha$ in \autoref{fig:voltron-framework} and in the training objective.

\subsection{Balancing Reconstruction \& Generation}
\label{subsec:voltron-balancing}

The \Voltron{} learning objective trades off language-conditioned \textit{reconstruction} and visually-grounded \textit{language generation} to shape the features captured by the encoder's learned representation. The reconstruction objective prioritizes low-level spatial information conducive to filling in missing textures, colors, or edges; likewise, the generation objective captures higher-level semantic information, encouraging the encoder to encode features that are predictive of the language caption. We make this tradeoff explicit by minimizing the following loss, characterized by the parameter $\alpha \in [0, 1]$:

\smallskip
\begin{align*}
    \mathcal{L}(\theta) &= \mathcal{L}_\text{reconstruct}(\theta) + \mathcal{L}_\text{generate}(\theta) \\
                        &=
        \begin{cases}
            \text{MSE}(v_\text{masked}, \text{R}_\theta(\text{E}_\theta(v_\text{visible}, c))) & \text{if } z = 0 \\
            \text{MSE}(v_\text{masked}, \text{R}_\theta(\text{E}_\theta(v_\text{visible}, \texttt{<NULL>}))) & \text{if } z = 1 \\
            \quad + \hspace{0.1em} \text{ NLL}(c, \text{G}_\theta(\text{E}_\theta(v_\text{visible}, \texttt{<NULL>}))) 
        \end{cases} \\
    \text{and} & \hspace{0.5em} z \sim \text{Bernoulli}(\alpha)
\end{align*}
\smallskip

For each example $(v, c)$ seen at training, we draw $z \sim \text{Bernoulli}(\alpha)$: with $z = 0$ we \textit{condition} on the original language utterance, while with $z = 1$, we \textit{generate} the original language utterance, conditioning the encoder on the \texttt{<NULL>} token. We limit our exploration in this work to at most two frame contexts $k = 2$ due to computational cost; even four frame contexts exceed the memory on the compute available to us. In selecting the two frame contexts, we sample at least five frames from each video clip in our dataset (with random intervals between). We enforce a heuristic such that the first frame in each dual-frame context comes from the first 20\% of the clip, with the other frame appearing in the remaining 80\%. 

Driven by the hypothesis that different values of $\alpha$ and frame-contexts $k$ shape the balance of low-level and high-level features in our representations, we evaluate three different instantiations of the \Voltron{} framework (as mentioned in \autoref{sec:introduction}): 
\begin{itemize}
    \item \VCond{}: $\alpha = 0$, $k = 1$ \textit{single-frame conditioning}.
    \item \VDual{}: $\alpha = 0$, $k = 2$ \textit{dual-frame conditioning}; a context-aware model identical to \VCond{} but trained on dual-frame pairs (initial frame, random subsequent frame).
    \item \VGen{}: $\alpha = 0.5$, $k = 2$; condition \textit{and} generate with equal probability, trained on dual-frame contexts as above.
\end{itemize}
Note that we \textit{do not evaluate} $\alpha = 1$; we find through preliminary experiments that some language-conditioning is always helpful.

\section{Implementation \& Reproducibility}
\label{sec:implementation-reproducibility}


In addition to our framework, a core contribution of this work is a comprehensive set of controlled experiments. To do this, we \textit{reimplement} both MVP and R3M using code released by the authors, controlling for the pretraining data (at the level of the individual frames seen per epoch) and model capacity.

\smallskip

\noindent \textbf{Baselines -- Preliminaries.} Throughout this work, we have mentioned both MVP and R3M in terms of their tradeoffs; here, we make their pretraining objectives explicit. Both prior approaches use video datasets, but only learn \textit{single-frame encoders}, choosing to use the video structure in different ways (detailed below). Of the two approaches, we note that only R3M uses language supervision.

MVP follows a masked autoencoding backbone, similar to that depicted in \autoref{fig:voltron-framework} (without language conditioning). MVP does not offer any special consideration to the temporal structure of videos, instead treating each frame in the dataset as as standalone input. Given a single frame, MVP masks out regions subject to a fixed mask ratio $\gamma$ (same as in \Voltron{}), encoding the visible context with a Transformer encoder, then attempting to reconstruct the missing context with a separate Transformer decoder -- also using mean-squared error for reconstruction. 

R3M is different in that it does not contain a reconstruction component, instead combining \textit{two contrastive objectives} on top of a single-frame visual encoder -- time contrastive learning \citep{sermanet2018tcn} and image-language temporal alignment \citep{radford2021clip, nair2021lorel}. These objectives explicitly use the \textit{temporal} structure of videos. Given an encoding of a visual context, the time-contrastive objective seeks to maximize the score of encodings between frames close together in time (e.g., within a few frames of each other), contrasted against frames from the same video that are further away. R3M also \textit{uses language supervision}. Given a separate encoder that fuses a language caption with the encoding dual-frames contexts (consisting of an initial and subsequent frame) the image-language alignment objective attempts to assign scores that capture ``task progress:'' the score of a subsequent frame occurring later in a video subject to a language caption should be higher than the score of a frame occurring earlier. The two key differences between \Voltron{} and R3M are 1) using visual reconstruction as a dense objective vs. time contrastive learning, and 2) explicitly conditioning on or generating language in \Voltron{} vs. matching visual and language embeddings as a contrastive objective.

\smallskip

\noindent \textbf{Pretraining Dataset Construction.} For all models in this work, we use Something-Something-v2 \citep[Sth-Sth;][]{goyal2017sthsth} as our pretraining dataset, motivated by prior work \citep{shao2020concept2robot, chen2021generalizable, xiao2022mvp}. All models see the \textit{exact same image frames}. We extract 5 frames per video, per training epoch to ensure we are learning from multiple visual inputs of the same context and to facilitate R3M's time contrastive learning objective \citep{sermanet2018tcn}; we serialize the processed frames, and store index files with the video/frame indices per epoch.

\smallskip

\noindent \textbf{Data-Equivalent Reproductions.} Though prior works release trained model artifacts, they do not provide sufficient details for reproduction, such as the exact frames sampled from videos, preprocessing applied, or hardware/compute used. We thus reimplement MVP and R3M in a controlled setting on Sth-Sth using the released code from the original papers where possible and clarifying additional details with the authors directly as needed. We implement all models with a Vision Transformer (ViT) backbone and additionally implement R3M with a ResNet-50 backbone based on discussions with the authors of the original work. They suggested that there may be slight differences in the inductive bias of ResNets vs. Vision Transformers \citep{raghu2021vitsvscnns} that would be worth investigating. We use the ViT-Small/16 variant, with patch size $p \times p = 16 \times 16$ and a Transformer with 12 blocks, 6 attention heads per block, and hidden dimension $d = 384$ \citep{wightman2019timm}. We refer to our reproductions as ``R-MVP,'' ``R-R3M (ViT-S),'' and ``R-R3M (RN-50).'' 

We pretrain all models in this work on TPU v3-8 compute, generously granted to us by the TPU Research Cloud program (TRC). We run 400 epochs of training for all models with a batch size of 1024, each epoch comprised of a pass through 844K frames (168K clips in Sth-Sth, 5 frames per clip). We do not use dropout or data augmentation. All code and reproducibility details are in our open-source code repositories, linked from our project page.

\smallskip

\noindent \textbf{Additional Comparisons.} Though we lack the compute resources to train on models on the same scale data, we further contextualize our results by evaluating the official R3M and MVP models released in the original works. We note that the released R3M model uses the entirety of the Ego4D dataset \citep{grauman2022ego4d}, comprised of over 3000 hours of videos, spanning 3.6M individual clips (comprising \textbf{more than 20x} the data we use in this work). The released MVP also uses Ego4D, but add Sth-Sth, Epic-Kitchens, and more \citep{damen2018kitchens, shan2020hands}, while also scaling models up to 86M and 307M parameters, (\textbf{4-10x} the size of ViT-Small). We also evaluate OpenAI's CLIP model (ViT-Base) as a strong baseline that leverages language supervision. We refer to these models as \unfair{``R3M (Ego4D),''} \unfair{``MVP (EgoSoup),''} and \unfair{``CLIP (ViT-B),''} following naming conventions from the original work and denote them with \unfair{gray text} and dashed lines in plots.

\begin{table*}[!th]
    \caption{\textbf{Summary of Evaluation Suite \& Results.} While some of our evaluation domains use language input, grasp affordance prediction and single-task visuomotor control \textit{do not}. While \Voltron{} models obtain strong performance over \textit{all applications}, R-R3M and R-MVP exhibit variable performance depending on the application subset.}
    \vspace*{-2mm}
    \label{tab:evaluation-summary}
    \small
    \begin{tabular*}{\linewidth}{@{\extracolsep{\fill}}lllccc@{}}
        \toprule
                                                                              & \textbf{Input Format}                          & \textbf{Train Dataset Size}   & \textbf{Best Model}   & \textbf{Best Baseline}  \\ \midrule
        \textbf{Grasp} \autoref{subsec:grasp-affordance}                               & Single Frame                                   & 1470                          & \VCond{}     & R-MVP              \\
        \textbf{Referring Expressions} \autoref{subsec:refer-detection}                & Single Frame, \textbf{Language Expression}     & 259,839                       & \VCond{}     & R-R3M (ViT)        \\
        \textbf{Single-Task Control} \autoref{subsec:single-task-control}             & Frame History                                   & $n \in [5, 10, 25]$ Demos     & \VDual{}     & R-R3M (RN-50)      \\
        \textbf{Language-Conditioned Imitation} \autoref{subsec:instruction-following} & Frame History, \textbf{Language Instruction}   & 100 = 5 x 20 Demos            & \VDual{} / \VGen{} & R-R3M (ViT)   \\
        \textbf{Intent Scoring} \autoref{subsec:intent-scoring}                        & Frame History, \textbf{Language Intent}        & N/A (Zero-Shot)               & \VGen{}      & N/A \\ \bottomrule
    \end{tabular*}
    \vspace*{1mm}
\end{table*}

\smallskip

\noindent \textbf{\Voltron{} Architecture Details.} \Voltron{} follows the masked autoencoding pipeline detailed above, with simple extensions for incorporating language. We implement the \Voltron{} encoder $\text{E}_\theta$ by jointly embedding the language $u$ and visual inputs $v_\text{visible}$ with a Transformer \citep{vaswani2017attention}. We initialize language embeddings from DistilBERT \citep{sanh2019distilbert}, learning a separate linear projection into the encoder's embedding space, similar to R3M. For the visual reconstructor $\text{R}_\theta$ and language generator $\text{G}_\theta$, we use a separate Transformer with a small addition to enable language generation. In a standard MAE decoder, patches are generated independently, attending to all patch embeddings from the encoder. To enable generation, we append a causal (lower triangular) attention mask for preventing our language decoder from ``peeking'' at the future inputs to generate (visualized by the \textcolor{xRed}{red triangle} in \autoref{fig:voltron-framework}). This is akin to prefix language modeling \citep{raffel2019exploring}; all embeddings can attend to the visual inputs (as in a traditional MAE decoder), but language embeddings can only attend to the preceding language input. 

\Voltron{} uses a combination of different language objectives on top of the standard MAE pipeline, adding complexity. To help ensure stable and reliable training, we follow best practices from the NLP community and make a series of small changes to the Transformer architecture including: 1) switching the default LayerNorm to root-mean square normalization \citep{zhang2019rms, narang2021transformermods} (stability, no learned parameters), 2) switching from the default GELU to the more performant SwishGLU activation  \citep{shazeer2020glu, chowdhery2022palm} (performance), and 3) adopting LayerScale for scaling down the magnitude of each residual connection \citep{touvron2021deeper, karamcheti2021mistral} (prevents overflow). To ensure that any gains in evaluation performance stem from our insights around language-driven learning rather than this modified architecture, we run an ablation experiment in \autoref{sec:ablations-extensions}. We find that these changes do not change downstream evaluation results, but significantly improve training stability. We present further details, including a sketch of the implementation differnces in \autoref{appx-subsec:voltron-transformer}.

\smallskip

\noindent \textbf{Adapting Representations.} Unfortunately, there is not yet a standard for extracting representations from learned Vision Transformer encoders, especially for those trained via masked autoencoding. However, \citet{zhai2022vitscaling} suggest that multiheaded attention pooling \citep[MAP;][]{lee2018settransformer} is a strong and versatile approach. We choose to use MAP as the sole feature extraction approach in all our ViT experiments, finding it to \textit{universally improve performance for all ViT models}, relative to the ``default'' extraction approaches suggested in prior work. Notably, we find that just switching to MAP-based extraction over the procedure used in the original MVP work \textit{almost doubles success rate} on visuomotor control tasks; we provide results from this analysis in \autoref{appx-subsec:map-feature-extraction}. We note that we use MAP when evaluating \unfair{CLIP (ViT-Base/16)} and \unfair{MVP (EgoSoup)} for the fairest and strongest possible comparison.

\section{Evaluation Suite: Construction \& Results}
\label{sec:evaluation-suite-results}


We outline our evaluation suite (\autoref{tab:evaluation-summary}) comprised of five problem domains within robotics. Each evaluation consists of \textit{adaptation data} and \textit{evaluation metrics}. The adaptation data consists of visual input(s) (as RGB frames) and in some cases, language (e.g., an instruction for language-conditioned imitation). We evaluate representations from \Voltron{} and various baseline models by \textit{freezing the pretrained vision and language encoders}, instead \textit{adapting} evaluation-specific ``heads''(lightweight networks) on top of the extracted representations. We choose evaluations that represent domains that capture different types of understanding; in the following sections, we motivate the role of each application and provide experimental results.

\begin{figure}[!t]
    \centering
    \includegraphics[width=\linewidth]{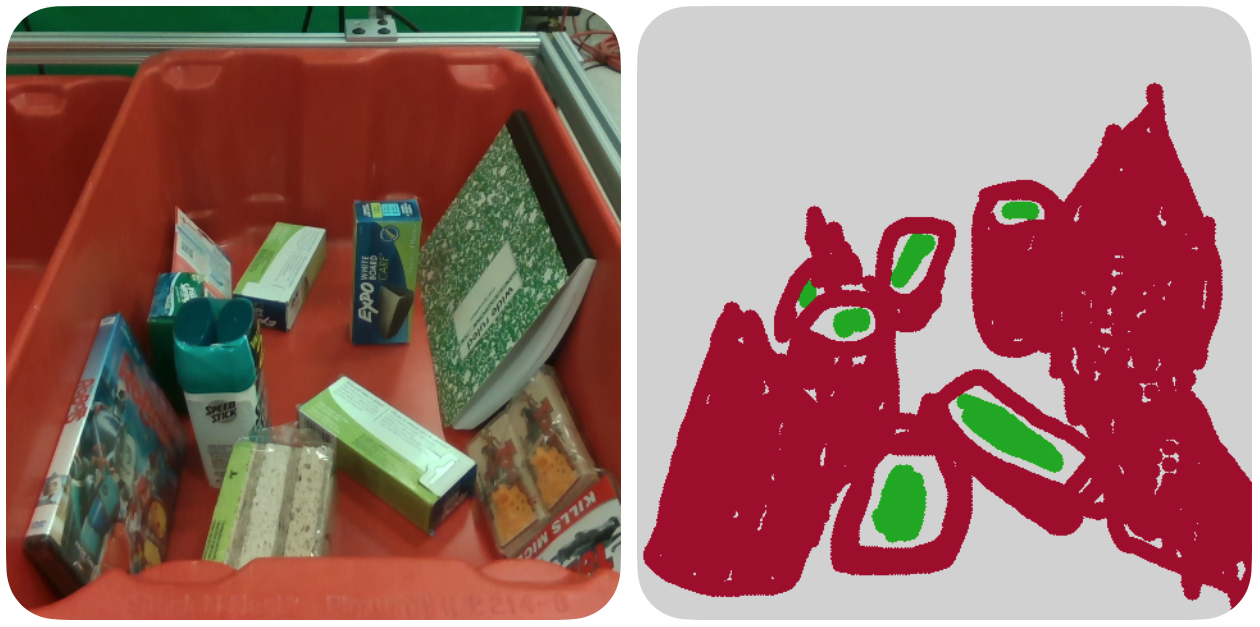}
    \vspace*{-5mm}
    \caption{\textbf{Grasp Affordance Prediction} \citep[ARC Grasping;][]{zeng2017arcgrasping}. Given objects in cluttered bins, segment the image corresponding to ``graspable'' (green), vs. ``non-graspable'' (red) regions; note that these regions are labeled for use with \textit{suction grippers}.}
    \label{fig:arc-grasping}
    \vspace*{-5mm}
\end{figure}

\begin{table}[!b]
    \caption{\textbf{Results on Grasp Affordance Prediction.} We report average precision at various confidence intervals following the original procedure described in \citet{zeng2017arcgrasping}.}
    \label{tab:arc-grasping-results}
    \vspace*{-2mm}
    \small
    \centering
    \begin{tabular*}{\linewidth}{@{\extracolsep{\fill}}lcccc@{}}
        \toprule
        \textbf{}                   & \textbf{Architecture} & \textbf{Top-1} & \textbf{Top 1\%} & \textbf{Top 5\%} \\ \midrule
        R-R3M                       & ViT-S      & \textit{40.38} & \textit{40.55}   & \textit{28.66}   \\
        R-MVP                       & ViT-S      & 72.94          & 61.47            & 39.77            \\
        \VCond{} [Ours] & ViT-S      & 85.15          & \textbf{80.71}   & 47.45            \\ \midrule
        \VCond{} [Ours] & ViT-B      & \textbf{90.00} & \textbf{82.44}   & \textbf{62.33}   \\ \midrule
        \unfair{CLIP}               & ViT-B      & \textit{43.20} & \textit{44.11}   & \textit{29.66}   \\
        \unfair{MVP (EgoSoup)}      & ViT-B      & 77.49          & 72.87            & 51.28            \\ \bottomrule
    \end{tabular*}
\end{table}

\begin{figure*}[t]
    \centering
    \includegraphics[width=\linewidth]{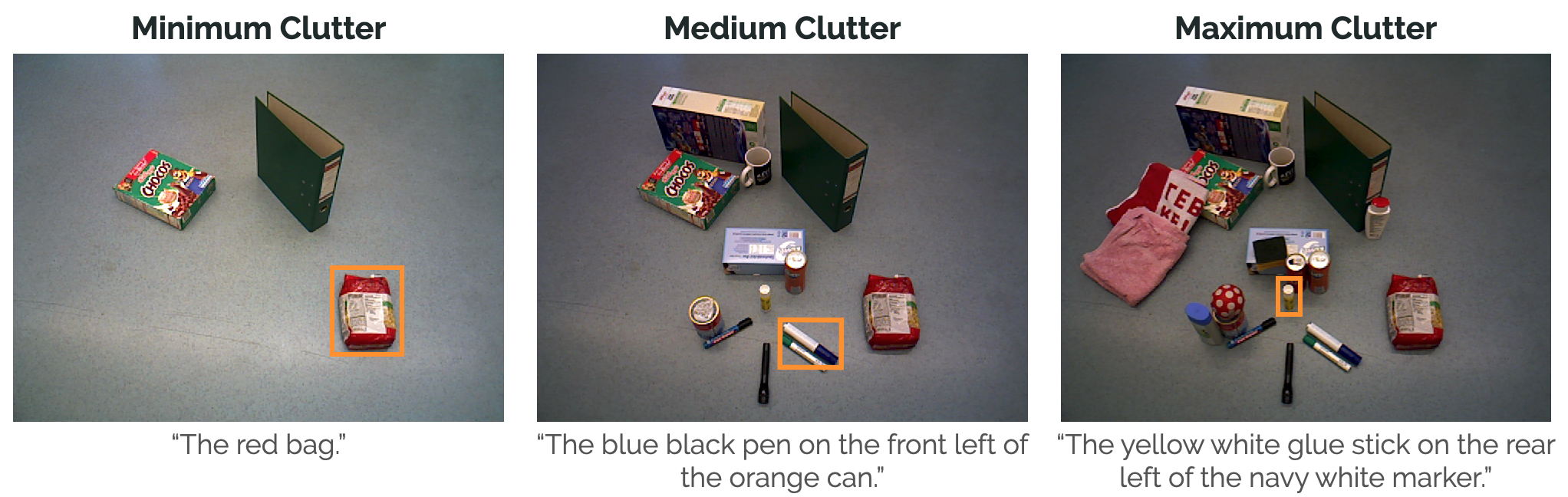}
    \vspace*{-5mm}
    \caption{\textbf{Referring Expression Grounding (Object Detection) from the OCID-Ref Dataset} \citep{wang2021ocidref}. Given a referring expression in natural language, the goal is to predict the bounding box coordinates around the respective object. An important feature of OCID-Ref are the various dataset splits, corresponding to three increasing amounts of clutter, depicted left-to-right.}
    \label{fig:ocid-ref}
\end{figure*}

\begin{table*}[hb]
    \vspace*{2mm}
    \caption{\textbf{Results on Referring Expression Grounding.} We report average precision @ 0.25 IoU following \citet{wang2021ocidref} (OCID-Ref). This is a \textit{language-conditioned} task; across various clutter levels, \Voltron{} models are substantially more performant than baselines, as well as models trained on more data and with alternative language supervision (e.g., CLIP).}
    \label{tab:ocid-ref}
    \vspace*{-2mm}
    \small
    \centering
    \begin{tabular*}{\linewidth}{@{\extracolsep{\fill}}lccccc@{}}
        \toprule
        \textbf{}                           & \textbf{Architecture} & \textbf{Total} & \textbf{Minimum Clutter} & \textbf{Medium Clutter} & \textbf{Maximum Clutter} \\ \midrule
        R-R3M                               & ViT-S         & 63.30          & 63.87       & 68.34 & 55.33  \\
        R-MVP + DistilBERT                  & ViT-S         & 49.58          & 50.98                & 53.83                & 41.94                           \\
        \VCond{} [Ours]         & ViT-S         & \textbf{89.38} & \textbf{85.88}       & \textbf{95.39}       & \textbf{89.12}                  \\ \midrule
        \VCond{} [Ours]         & ViT-B         & \textbf{90.77} & \textbf{87.56}       & \textbf{96.58}       & \textbf{90.17}                  \\ \midrule
        \unfair{CLIP}                       & ViT-B         & 68.35          & 67.01                & 76.61                & 60.33                           \\
        \unfair{MVP (EgoSoup) + DistilBERT} & ViT-B         & 49.25          & 51.46                & 52.15                & 40.50                           \\ \bottomrule
    \end{tabular*}
\end{table*}

\subsection{Grasp Affordance Prediction}
\label{subsec:grasp-affordance}

We consider the problem of grasp affordance prediction: given an image of a set of objects (e.g., on a cluttered workspace), predict a dense segmentation mask corresponding to ``graspable'' and ``non-graspable'' locations for a suction-based gripper. 

\smallskip

\noindent \textbf{Motivation.} Grasp affordance prediction from visual input is a foundational task in robot learning, and is often a key component of many modular systems \citep{bohg2013datadrivengrasping, correll2016analysis}. Including this evaluation allows us to probe the low-level spatial features retained by various representations.

\smallskip

\noindent \textbf{Evaluation Details.} We specifically consider the problem as formulated in the Amazon Robotics Challenge Grasping Dataset (ARC-Grasping) introduced by \citet{zeng2017arcgrasping}. We choose this dataset over alternatives as it is readily available and consists of 1800+ images of multiple real-world objects in cluttered bins (\autoref{fig:arc-grasping}; left). We focus on the RGB-only, suction-grasping split of the dataset. We implement models for grasp affordance prediction following recent work on semantic segmentation with Transformers \citep{zheng2021setr, strudel2021segmenter, bao2022beit}, specifically by introducing a Progressive Upsampling (SETR-PUP) head on top of our frozen visual features. We omit results from all ResNet models -- R-R3M (RN-50) and \unfair{R3M (Ego4D)}; unfortunately, training with simple PUP-style on the final ResNet-50 $7 \times 7$ spatial grid did not converge, possibly indicating a need for more complex architectures with significant added parameters (beyond the scope of this work). As this task only takes a single frame as input, we do not evaluate \VDual{} and \VGen{}. Following the original work, we report average precision at various confidences: Top-1 precision, Top-1\% precision, and Top-5\% precision. We select models via 5-fold cross validation. This task \textit{does not have a language component}. We provide additional details around the adaptation procedure in \autoref{appx:adaptation-for-evaluation} and the open-source code repositories.

\medskip

\noindent \textbf{Experimental Results.} Looking at \autoref{tab:arc-grasping-results}, representations from MVP and \Voltron{} models perform well across the board, while contrastive representations (e.g., from CLIP and R-R3M) perform quite poorly. Interestingly, \VCond{} outperforms R-MVP and \unfair{MVP (EgoSoup)} on this task, \textit{despite the absence of language input}, demonstrating that language supervision during pretraining can improve low-level feature learning, even relative to larger-scale models trained on much more data.

\begin{figure*}[t]
    \centering
    \includegraphics[width=\linewidth]{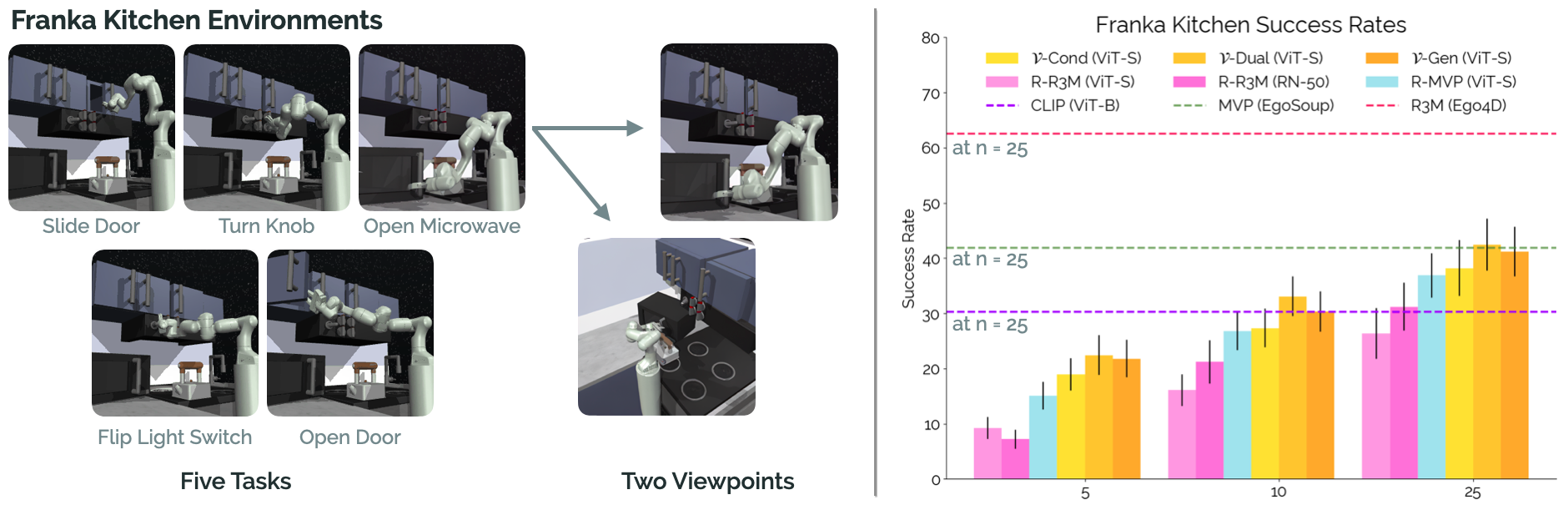}
    \vspace*{-5mm}
    \caption{\textbf{Franka Kitchen -- Single-Task Visuomotor Control Results}. Visualization of the Franka Kitchen evaluation environments, comprised of five unique tasks, with two camera viewpoints \textbf{[Left]}. Results (success rate for each of $n$ demonstrations) for \Voltron{} and baselines, showing the benefit of language-driven learning (over 3 seeds) \textbf{[Right]}. In dashed lines (not directly comparable), we plot \unfair{CLIP (ViT-B)}, \unfair{MVP (EgoSoup)}, and \unfair{R3M (Ego4D)} trained with $n = 25$ demonstrations.}
    \label{fig:control-franka-kitchen}
\end{figure*}

\subsection{Referring Expression Grounding}
\label{subsec:refer-detection}

Given a cluttered scene and language expression, the goal is to predict a bounding box around an object (e.g., ``the blue black pen on the front left of the orange can'' in \autoref{fig:ocid-ref}; middle).

\smallskip

\noindent \textbf{Motivation.} Capturing object-centric priors and high-level semantics around properties such as color and spatial relationships is crucial across the entire robotics stack. More importantly, this is a \textit{language-conditioned} task, allowing us to evaluate the impact of pretraining with language supervision.

\smallskip

\noindent \textbf{Evaluation Details.} We use the OCID-Ref Dataset \citep{wang2021ocidref} grounded in scenes that are representative of robotics settings; other datasets such as RefCoCo \citep{yu2016refcoco} are grounded in more global scenes (e.g., multiple humans playing frisbee on a field) that are less informative for robot learning. OCID-Ref also provides splits based on the clutter level of the underlying scene, letting us further evaluate robustness. We regress bounding box coordinates directly from our frozen features using a shallow MLP. All approaches condition on language (see expressions in \autoref{fig:ocid-ref}), using the given language encoder where possible. This means using the multimodal encoder for \VCond{} and the default learned text encoder for CLIP or R3M. However, for approaches that only learn visual representations (e.g., MVP), we append pretrained language features from DistilBERT -- the same language model used to initialize \Voltron{}. We note again that we omit ResNet results; though this task did not require upsampling, we find trained models obtained no better than random performance, again indicating a need for a more sophisticated adaptation architecture (beyond the scope of this work). We report average precision at 0.25 IoU for each split following the evaluation procedure outlined in \citet{wang2021ocidref}. We provide additional details around the adaptation procedure in \autoref{appx:adaptation-for-evaluation} and the open-source code repositories.

\smallskip

\noindent \textbf{Experimental Results.} Results for each model across the various clutter splits are in \autoref{tab:ocid-ref}. \Voltron{} models are especially strong, vastly outperforming R-MVP by 40\% and R-R3M by over 25\% on all splits, showing that multimodal pretraining -- even just conditioning on language when optimizing for masked reconstruction -- can lead to substantial gains on downstream multimodal tasks. We isolate the massive performance gains of \Voltron{} models over prior work due to the multimodal encoder that learns \textit{fused} embeddings of vision and language, allowing language to shape the visual representations during pretraining. In contrast, R3M, and CLIP models learn \textit{independent} text encodings that are only fused post-hoc, during adaptation. This is even worse for MVP: these models need to learn to fuse their strong visual embeddings with the language embeddings from a completely different model (DistilBERT).

\subsection{Single-Task Visuomotor Control}
\label{subsec:single-task-control}

\noindent \textbf{Motivation.} Imitation learning for visuomotor control has been the de-facto evaluation for prior work \citep{parisi2022unsurprising, nair2022r3m, radosavovic2022mvp}, giving us the closest comparison to the evaluations used in MVP and R3M. This evaluation focuses on \textit{sample-efficient generalization}, measuring how well visual representations help in learning policies from limited demonstrations $n \in \{5, 10, 25\}$. This evaluation takes place in simulation.

\smallskip

\noindent \textbf{Evaluation Details.} We look at policy learning in the Franka Kitchen simulation environments as defined by \citet{nair2022r3m}. This domain consists of 5 tasks, with 2 distinct camera viewpoints (\autoref{fig:control-franka-kitchen}). We learn shallow MLP policy heads via behavioral cloning that predict 9-DoF joint velocities (7 joints, 2 gripper) from our (frozen) visual features and proprioceptive state. We follow the R3M evaluation, reporting average success rates for each setting with $n$ demonstrations across the 5 tasks, 2 viewpoints, and 3 random seeds. We train separate policies per task, with \textit{no language conditioning} -- using the exact code provided by \citet{nair2022r3m}. Additional details are in \autoref{appx:adaptation-for-evaluation} and the open-source code.

\smallskip

\noindent \textbf{Experimental Results.} Most approaches perform similarly across the various number of training demonstrations (\autoref{fig:control-franka-kitchen}; right). However, we see some promising trends; \Voltron{} models perform better than both baselines, with approaches that learn from multiple frame contexts \VDual{} and \VGen{} showing \textit{significant} improvements over single-frame approaches. Yet, the absolute success rates are low; learning for control is difficult, and while good visual representations can help, learning closed-loop policies from limited data remains an open challenge.

\begin{figure*}[t]
    \centering
    \includegraphics[width=\linewidth]{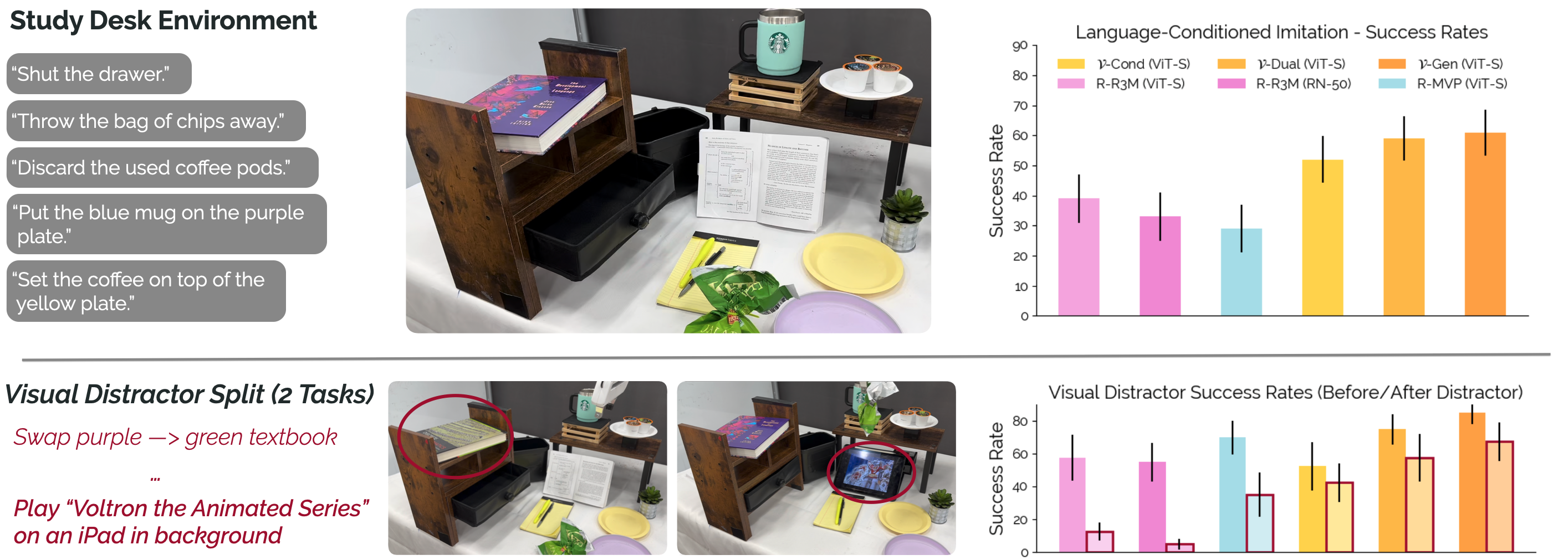}
    \vspace*{-5mm}
    \caption{\textbf{Real-World Language-Conditioned Imitation Learning Results}. The real-world ``Study Desk'' environment, with sample language instructions corresponding to the five behaviors we evaluate. \textbf{[Top]} The challenging \textit{visual distractor} split for evaluating robustness to novel distractors, ranging from simple color swapping of background objects (e.g., purple to green textbook), to more drastic changes such as playing a clip from \textit{``Voltron -- the Animated Series''} in the background \textbf{[Bottom]}.}
    \label{fig:instruct-real-robot}
\end{figure*}

\subsection{Language-Conditioned Imitation (Real)}
\label{subsec:instruction-following}

Given a dataset of language instructions (e.g. ``throw the bag of chips away'') paired with demonstrations (on a real robot in a real-world tabletop setting), learn an instruction following policy via behavioral cloning. \autoref{fig:instruct-real-robot} depicts the real-world environment.

\smallskip
\noindent \textbf{Motivation.} A large body of work looks at learning language-conditioned policies for human-robot collaborative settings \citep{arumugam2017accurately,stepputtis2020lcil,lynch2020grounding,karamcheti2021lila,ahn2022saycan}. This evaluation gets at the robustness and reliability of learned representations, with the goal of validating different approaches in real-robot settings.

\smallskip
\noindent \textbf{Evaluation Details.} We construct a ``study desk'' environment (\autoref{fig:instruct-real-robot}) with five prototypical ``tasks'': 1) closing the drawer, 2) throwing the green bag of chips in the trash can, 3) discarding the used coffee pods, 4) moving the cyan coffee mug to the purple plate, and 5) moving the same mug to the yellow plate. For each task, we collect 20 teleoperated demonstrations at 10 Hz, randomly resetting the scene between episodes. We adopt the keyframe-based action space proposed in \citet{james2022qattention} for learning. This approach heuristically breaks a demonstration into 4-5 ``waypoints'' (end-effector poses) that are used as action targets during behavior cloning; during policy execution, we plan min-jerk trajectories from the current position to the predicted waypoint, feeding the subsequent state and visual observation back to our policy \citep{james2022coarse2fineq, shridhar2022peract}. To collect diverse instructions, we prompt ChatGPT \citep[version dated Jan 9th, 2023;][]{openai2022chatgptjan9th} with simple task descriptions, asking it to generate diverse language instructions, collecting 25 utterances total (20 train, 5 held-out) per task.\footnote{ChatGPT Prompt (additional details and generated instructions on project page): \textit{I'm trying to train a robot assistant that can follow diverse language instructions. One task requires moving an empty chip bag (a green bag of those jalapeno chips) to the garbage. Can you generate 25 natural-sounding instructions (e.g., ``throw away the chips'')?}} We parameterize our policy similarly to \autoref{subsec:single-task-control}, adding a shallow MLP on top of the extracted (frozen) visual representations \citep{misra2017mapping}. This task is \textit{language-conditioned}; as in OCID-Ref, we use the given language encoders for each approach where possible, appending DistilBERT features to pure visual representations otherwise. We report success rates with partial credit -- 0.25 points for achieving each of the following ``milestones'': reaching an object, interacting with it, transporting it, and completing the task. We provide additional details in \autoref{appx:adaptation-for-evaluation}, and include videos of policy rollouts on the project page.

\smallskip

\noindent \textbf{Experimental Results.} Looking at success rates of the various representations (\autoref{fig:instruct-real-robot}; top right) we see an exaggerated version of the trends exhibited in the single-task control setting; \Voltron{} models obtain an extra boost in performance across the board given that this task is language-conditioned, highlighting the strength of its fused representations. Similarly, R-R3M models exhibit the next best performance. Due to time and shared resource constraints, we do not run out \unfair{MVP (EgoSoup)}, \unfair{R3M (Ego4D)}, or \unfair{CLIP (ViT-B/16)}, though we expect similar trends as in the last evaluation.

\begin{figure*}[t]
    \centering
    \includegraphics[width=\linewidth]{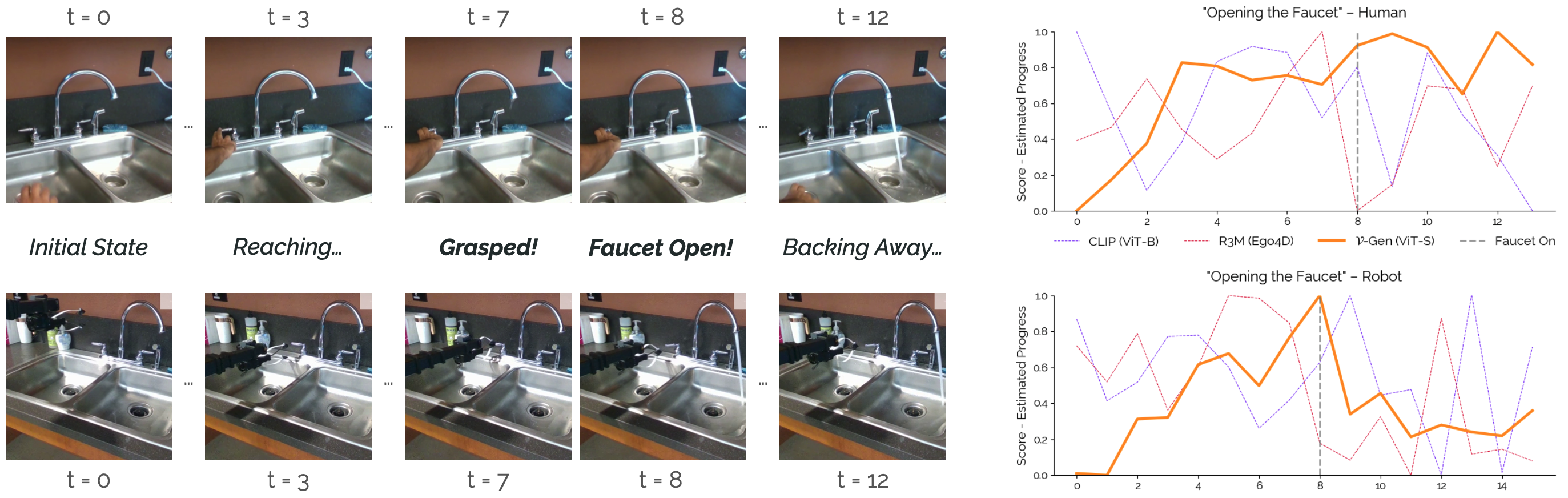}
    \vspace*{-5mm}
    \caption{\textbf{Qualitative Zero-Shot Intent Scoring Results.} Given a pair of videos from the WHiRL dataset \citep{bahl2022whirl} of a human and robot performing a task, we evaluate the ability of \VGen{}, R3M (from \citet{nair2022r3m}) and CLIP in scoring various frames subject to the utterance ``opening the faucet.'' While CLIP and R3M produce extremely noisy scores, \VGen{} is \textit{calibrated}, successfully tracking progress over time -- both for the human user, \textit{as well as for the robot}.}
    \label{fig:intent-scoring}
\end{figure*}

\subsection{Qualitative: Zero-Shot Intent Scoring}
\label{subsec:intent-scoring}

We perform a qualitative evaluation for the problem of language-based \textit{intent scoring}; given a language expression describing an intent or behavior (e.g., ``opening the faucet'') and a corresponding video (that may or may not show the described behavior), predict an ``alignment score'' for each frame of a video. This alignment score should capture how well the current visual context matches the described behavior -- ideally reflecting calibrated confidence over time (an example language/video is shown in \autoref{fig:intent-scoring}; left).

\smallskip
\noindent \textbf{Motivation.} This evaluation is motivated by two active areas of research: reward learning from language and demonstrations \citep{smith2020avid, shao2020concept2robot, chen2021generalizable, bahl2022whirl}, and belief modeling for human-robot collaboration \citep{hoffman2007cost, hauser2012recognition, bandyopadhyay2013intention} This evaluation probes for the ability to reason over intents and visual behaviors \textit{jointly, without the need for additional data}.

\smallskip
\noindent \textbf{Evaluation Details.} This is a qualitative evaluation that focuses on measuring how well existing approaches ``track'' progress conditioned on a language intent over time. Doing this zero-shot means that we can only evaluate models that can produce alignment scores given language and visual context: 1) \unfair{CLIP (ViT-B/16)} through cosine similarity of learned vision and text representations, 2) \unfair{R3M (Ego4D)} through the ``video-language alignment'' head, and 3) our \VGen{} model  (by measuring the likelihood of a given language utterance conditioned on visual context under the language generator). Given a video of an agent performing some behavior described in language (e.g., ``opening the faucet''), we estimate and plot scores under each model across a sequence of video frames. We use videos from WHiRL \citep{bahl2022whirl} of humans and robots performing the same tasks from different views; we choose to evaluate intent scoring for both agents to better capture the robustness and transfer potential for these approaches in similar real-world settings. 

\smallskip

\noindent \textbf{Experimental Results.} The two curves in \autoref{fig:intent-scoring} show the predicted scores over time for the language intent ``opening the faucet.'' Even though it has never been trained for this task, we find that \VGen{} is able to coherently predict not only the exact frames corresponding to ``keypoints'' in each video (e.g., touching the handle, observing when the water starts running), but is also capable of measuring \textit{partial progress} -- akin to a shaped, dense reward; however, both \unfair{R3M (Ego4D)} and \unfair{CLIP (ViT-B/16)} fail at this task, predicting random scores with high variance across sequential time steps. Note that the intent scores are not perfect; after turning the faucet on for the human video, predicted scores remain high, while for the robot, the scores taper off. It is not clear why this happens, but given a small amount of adaptation data, one could ensure consistent behavior. We provide more examples from WHiRL in \autoref{appx-subsec:additional-intent-scoring}, and additional evaluation details in \autoref{appx:adaptation-for-evaluation}.

\section{Ablations, Extensions, \& Further Analysis}
\label{sec:ablations-extensions}


The comparative results across the various evaluation problem domains paint \Voltron{}'s language-driven representations in a favorable light relative to MVP and R3M baselines. Yet, there remain key questions that we address in this section: is language supervision actually driving these results? Why generative language modeling over masked language modeling? Will \Voltron{} scale?

\smallskip
\noindent \textbf{Ablation: The Impact of Language Supervision.} The second row of \autoref{tab:ablation-results} shows a subset of evaluation results across three different problem domains when training a ``no-language'' variant of the \VCond{} architecture -- this variant is in essence an alternate version of a masked autoencoder that uses the small architecture modifications we added for training stability in \autoref{sec:implementation-reproducibility}. As such, it also serves as an \textit{architecture ablation} when compared to the R-MVP results, enabling us to isolate the impact of the small stability modifications described in \autoref{sec:implementation-reproducibility}. Indeed, the results confirm our hypotheses: first, removing language results in a definitive drop in performance across all evaluation applications. Second, the respective results for each evaluation application are on par with the corresponding results for the R-MVP model, demonstrating that the performance of \Voltron{} models does not stem from the architecture. We delve further into this ablation in \autoref{appx-subsec:impact-of-language}.

\smallskip

\noindent \textbf{Ablation: Generative vs. Masked Language Modeling.} Looking at the \Voltron{} objective, a natural question to ask is why we chose \textit{language generation} over \textit{masked language modeling}. Furthermore, recent and concurrent work propose learning multimodal masked autoencoders (M3AE) both within and outside of robotics \citep{geng2022m3ae,liu2022instructrl}, showing promising results in learning visual representations for image classification tasks, amongst others. To assess the differences, we choose to reproduce the M3AE model in a manner similar to our reproduction of MVP and R3M; we keep the same Something-Something-v2 pretraining data, adopting the exact procedure described in \citet{geng2022m3ae}, then evaluating the resulting representations on the same subset of evaluation domains as in the prior ablation (third row of \autoref{tab:ablation-results}). Surprisingly, we see drastic drops in performance \textit{across the board}. Looking at the pretraining curves, we identify a possible reason for this failure: in optimizing M3AE on Sth-Sth, we see the language modeling loss go to zero almost \textit{immediately}, leading to overfitting. A possible explanation is that the masked language modeling conditioned on visual contexts in datasets annotated with \textit{short, predictable narrations} leads to degenerate representations, while generative language modeling is not susceptible to the same types of collapse; looking at ways to mitigate this seems like a promising direction for future work. Explicit details around pretraining and evaluating R-M3AE, with an in-depth discussion are in \autoref{appx-subsec:m3ae-results}.

\smallskip

\noindent \textbf{Extension: Scaling Up.} Prior approaches have shown gains in scaling model capacity; here, we present preliminary evidence that \Voltron{} models behave similarly. For each evaluation in \autoref{sec:evaluation-suite-results}, we evaluate a ViT-Base variant of \VCond{} (86M parameters vs. the 22M in the ViT-Small). We see universal improvement: Top-5\% precision for grasping (\autoref{tab:arc-grasping-results}; middle row) increases by 15\%, expression grounding accuracy improves (\autoref{tab:ocid-ref}; middle row), as does performance on control.

\smallskip

\noindent \textbf{Extension: Robustness to Real-World Distractors.} Factors such as lighting conditions, time of day, and accidental environment perturbations (e.g., a colleague knocking over the camera) can have a profound impact on performance of robotic systems, especially if learned representations are not robust. We run a limited ``robustness'' evaluation after training language-conditioned policies from the demonstrations described in \autoref{subsec:instruction-following}. Success rates before and after introducing visual distractors for two of the ``meta-tasks'' are in \autoref{fig:instruct-real-robot} (bottom right).\footnote{We try five distractors spanning simple changes such as swapping the purple textbook in the background for a green one, to more extreme distractors such as playing a clip from \textcolor{xOrange}{\textit{``Voltron, the Animated Series''}} on a tablet in the middle of the workspace. Videos are on the project page.} We find that \Voltron{} and R-MVP models are robust to even the most extreme distractors -- seemingly a benefit of per-patch masking coupled with MAP-based extraction.

\section{Discussion \& Conclusion}
\label{sec:discussion}

\begin{table}[!t]
    \caption{\textbf{Ablation Experiments.} We select a subset of evaluations from \autoref{sec:evaluation-suite-results} -- grasp affordance prediction, referring expression grounding, and single-task visuomotor control.}
    \label{tab:ablation-results}
    \vspace*{-2mm}
    \small
    \centering
    \begin{tabular*}{\linewidth}{@{\extracolsep{\fill}}lccc@{}}
        \toprule
                                                    & \textbf{Grasp}     & \textbf{Refer}   & \textbf{Imitate}  \\
                                                    & PR @ Top-1\%       & Total Accuracy   & (n = 25)          \\ \midrule
        $\mathcal{V}$ + Lang [Ours]                                     & 80.71             & 89.38             & 38.2 $\pm$ 5.09           \\ \midrule
        No-Language $\boldsymbol{\downarrow}$                           & 65.83             & \textit{53.44}    & 33.1 $\pm$ 4.79           \\
        R-M3AE \textcolor{red}{$\boldsymbol{\downarrow \downarrow}$}    & \textit{52.79}    & \textit{51.61}    & \textit{24.0 $\pm$ 4.21}  \\ \bottomrule
    \end{tabular*}
\end{table}


We propose \Voltron{}, a framework for language-driven representation learning that balances \textit{conditioning} and \textit{generation} to shape the balance of low and high-level features captured. We introduce an evaluation suite spanning five diverse problems within robotics for holistically evaluating visual representations. Through controlled experiments and ablations, we validate the strengths of our representations; across all evaluation tasks, \Voltron{} models that balance language conditioning and generation strictly outperform prior approaches such as R3M and MVP, and in many cases show performance competitive with or exceeding that of approaches that use orders of magnitude more data or more expressive models. 

Yet, while language is a pivotal source of supervision, there are still key questions to answer. Why is language-based pretraining helpful on tasks that have nothing to do with language? Why not try to learn \textit{one model} that can encode both low-level and high-level features, without tradeoffs? While there is not a silver bullet \textit{yet} we hope that future work takes a deep, grounded look at these questions, identifying what existing representations capture -- and more importantly, what they miss. Our hope is that \Voltron{} serves as a starting point; a flexible, unified framework for future improvements in visual representation learning for robotics.

\begin{acks}
This work would not have been possible without the support of entire communities of students, engineers, and various domain experts; our gratitude cannot be understated. We would specifically like to thank Shyamal Buch, David Hall, Sasha Khazatsky, and John Thickstun for their invaluable advice and suggestions around pretraining and evaluation. We further thank Dilip Arumugam, Masha Itkina, Minae Kwon, Tyler Lum, Vivek Myers, and Karl Pertsch for their feedback on earlier drafts.

Toyota Research Institute (``TRI'') provided funds to support this work. This project was additionally supported by the Office of Naval Research (ONR). Parts of this research -- specifically model pretraining -- was supported with Cloud TPUs from Google's TPU Research Cloud (TRC). Siddharth Karamcheti is grateful to be supported by the Open Philanthropy Project AI Fellowship. Annie Chen is supported by the NSF Graduate Research Fellowship (NSF GRFP). Finally, we thank the members of the Stanford ILIAD, IRIS, and NLP groups for valuable discussions and their unwavering support. 
\end{acks}

\newpage
\bibliographystyle{x-acm-reference-format}
\bibliography{x-refdb}

\newpage
\onecolumn
\appendix

\section*{Overview}

\bigskip
\noindent In the appendices below, we provide additional details around the implementation, pretraining, and adaptation procedures described in the main text, in addition to delving deeper into various discussions. Finally, we add additional results and visualizations that further complement the findings from the main text.

\medskip

\noindent We provide open-source code for loading and using pretraining models, hosted links for our preprocessing splits (including the actual batches seen during training), and a separate, standalone open-source code repository for our evaluation suite. Our hope is that the evaluation suite especially is general and easy to use for downstream work on evaluating learned representations. The full manifest of resources are as follows:
\begin{itemize}
    \item Project Page (videos \& additional links): \url{https://sites.google.com/view/voltron-robotics}
    \item Open-Source Modeling Repository (pretraining code for \textit{all approaches}, loading models): \url{https://github.com/siddk/voltron-robotics}
    \item Open-Source Evaluation Suite (general API for evaluating on different problem domains): \url{https://github.com/siddk/voltron-evaluation} 
\end{itemize}
All model and automated evaluation code is in PyTorch; however, the evaluation code can be easily overridden to suit your needs.

\medskip

\noindent An overview of each appendix can be found below. We further indicate which parts of the appendices are best viewed here in the text or on the project page; for videos and visualizations, we highly recommend navigating to the latter.

\medskip
\noindent \rule{\linewidth}{0.5pt}
\medskip

\noindent \hyperref[appx:motivating-questions]{\autoref{appx:motivating-questions} -- \textit{Motivating Questions}} 

\begin{quote}
    \medskip
    We index a list of ``motivating'' questions that may arise from reading the main text and that we expand on further here (e.g., ``why only evaluate frozen representations''). Our answers here are \textit{direct}, and in many cases link to actual experiments further on in the appendices.
\end{quote}

\bigskip

\noindent \hyperref[appx:voltron-implementation]{\autoref{appx:voltron-implementation} -- \textit{\Voltron{} Implementation}}

\begin{quote}
    \medskip
    We provide code and other implementation details around the modifications to the Transformer architecture described in the Implementation and Reproducibility Section (see \autoref{sec:implementation-reproducibility}) of the main text, along with additional details around the released models and data artifacts from this work. The section is structured as follows:
    \medskip
    \begin{itemize}
        \item[] \hyperref[appx-subsec:voltron-transformer]{\autoref{appx-subsec:voltron-transformer} -- \textit{\Voltron{} Transformer Implementation}}
        \begin{quote}
            \medskip
            Side-by-side comparisons of the \Voltron{} and ``standard'' Vision Transformer blocks.
            \medskip
        \end{quote}
        \item[] \hyperref[appx-subsec:multimodal-block-details]{\autoref{appx-subsec:multimodal-block-details} -- \textit{Jointly Processing Vision \& Language}}
        \begin{quote}
            \medskip
            Additional details around encoding multimodal inputs (e.g., position encoding, modality tokens, etc.).
            \medskip
        \end{quote}
        \item[] \hyperref[appx-subsec:pretraining-curves]{\autoref{appx-subsec:pretraining-curves} -- \textit{Pretraining Curves}}
        \begin{quote}
            \medskip
            \Voltron{} pretraining loss curves (reconstruction error, language modeling error) over training; useful for characterizing the behavior
            of downstream models (and the trade-offs between the losses).
            \medskip
        \end{quote}
        \item[] \hyperref[appx-subsec:released-curves]{\autoref{appx-subsec:released-curves} -- \textit{Index of Released Artifacts}}
        \begin{quote}
            \medskip
            We release pretrained \Voltron{} models -- \VCond{}, \VDual{}, \VGen{} -- in addition to intermediate checkpoints to facilitate future work. We also release the larger \VCond{} model (ViT-Base).
        \end{quote}
    \end{itemize}
\end{quote}

\bigskip

\noindent \hyperref[appx:results-visualizations]{\autoref{appx:results-visualizations} -- \textit{Additional Results \& Visualization}}

\begin{quote}
    \medskip
    We report additional results and visualizations from experiments mentioned in the main text, as well as other experiments that further support our conclusions.
    \medskip
    \begin{itemize}
        \item[] \hyperref[appx-subsec:impact-of-language]{\autoref{appx-subsec:impact-of-language} -- \textit{Analysis: Impact of Language-Conditioning on Reconstruction Loss}}
        \begin{quote}
            \medskip
            We revisit the language vs. no-language ablation from the main text, looking at pretraining curves to help explain why language is so helpful as a supervision signal. We find that language-conditioning significantly lowers reconstruction loss, allowing models to pick up on more low-level features.
            \medskip
        \end{quote}
        \item[] \hyperref[appx-subsec:m3ae-results]{\autoref{appx-subsec:m3ae-results} -- \textit{Analysis: Generative vs. Masked Language Modeling}}
        \begin{quote}
            \medskip
            We look further at the masked language modeling ablation from the main text, via the reproduction of Multimodal Masked Autoencoders \citep[M3AE;][]{geng2022m3ae}. We find in the pretraining curves high evidence of overfitting with masked models early in training, impacting the learned representations.
            \medskip
        \end{quote}
        \newpage
        \item[] \hyperref[appx-subsec:adroit-single-task-results]{\autoref{appx-subsec:adroit-single-task-results} -- \textit{Results: Adroit Visuomotor Control}}
        \begin{quote}
            \medskip
            We present results on the Adroit Visumotor Control environments from \citet{nair2022r3m}, finding that while language is again superior, \textit{higher-level features perform better}. This is preliminary evidence that even for individual evaluation domains (e.g., single-task visuomotor control), there is no silver bullet; different types of representations perform differently.
            \medskip
        \end{quote}
        \item[] \hyperref[appx-subsec:real-robot-rollouts]{\autoref{appx-subsec:real-robot-rollouts} -- \textit{Qualitative: Real-Robot Language-Conditioned Policy Rollouts}}
        \begin{quote}
            \medskip
            Visualizations of real-world policy rollouts from the various representation learning approaches.
            \medskip
        \end{quote}
        \item[] \hyperref[appx-subsec:additional-intent-scoring]{\autoref{appx-subsec:additional-intent-scoring} -- \textit{Qualitative: Additional Intent Scoring Visualizations}}
        \begin{quote}
            \medskip
            Additional intent scoring visualizations using videos from the WHiRL dataset \citep{bahl2022whirl}.
        \end{quote}
    \end{itemize}
\end{quote}

\bigskip

\noindent \hyperref[appx:full-implementation-reproducibility]{\autoref{appx:full-implementation-reproducibility} -- \textit{Data-Equivalent Reproductions \& Reproducibility}}

\begin{quote}
    \medskip
    We add additional discussion around the reproductions of MVP and R3M on the Something-Something-v2 dataset:
    \begin{itemize}
        \item[] \hyperref[appx-subsec:preprocessing-discussion]{\autoref{appx-subsec:preprocessing-discussion} -- \textit{Additional Preprocessing Discussion}}
        \begin{quote}
            \medskip
            Additional discussion of how we preprocess Something-Something-v2 \citep[Sth-Sth;][]{goyal2017sthsth} for pretraining, with a comparison of how prior work such as MVP source and process pretraining data.
            \medskip
        \end{quote}
        \item[] \hyperref[appx-subsec:map-feature-extraction]{\autoref{appx-subsec:map-feature-extraction} -- \textit{Multiheaded Attention Pooling -- Feature Extraction}}
        \begin{quote}
            \medskip
            Detailed explanation of the Multiheaded Attention Pooling \citep[MAP;][]{lee2018settransformer} feature extraction strategy, with analysis and results comparing to alternative methods.
        \end{quote}
    \end{itemize}
\end{quote}

\bigskip

\noindent \hyperref[appx:adaptation-for-evaluation]{\autoref{appx:adaptation-for-evaluation} -- \textit{Adapting Representations for Evaluation}}

\begin{quote}
    \medskip
    We provide further descriptions of the adaptation pipeline for each of the five evaluation domains.
\end{quote}

\newpage

\section{Motivating Questions}
\label{appx:motivating-questions}


\bigskip
\noindent \textbf{Q1.} \textit{From the results, some \Voltron{} models outperform larger models such as MVP-Base trained on significantly more data, even on tasks that do not necessarily need language information. How do you make sense of this?}

\medskip

We find that in many of our evaluation domains, especially domains with episodic tasks such as single-task and language-conditioned imitation learning, it is important to discern differences \textit{across frames in the same overall visual context}, or otherwise pay attention to small visual distinctions. Looking at the original MVP work \citep{radosavovic2022mvp}, we see that the original pretraining datasets are compiled by sampling frames from various video datasets \textit{once, in a single-step procedure}, at low sampling rates. For many datasets (such as Sth-Sth and Ego4D), this means only seeing 1-2 frames per video clip in total during training.

\medskip

In contrast, when we sample data from Sth-Sth, we ensure to sample \textit{at least 5 frames per clip, per epoch}; while the aggregate amount of diverse contexts is much lower than in the original MVP work, seeing multiple frames per context seems to significantly help learning, \textit{and not just for \Voltron{} models}! On the tasks where \Voltron{} models outperform \unfair{MVP (EgoSoup)} (with a larger ViT-Base encoder), we also see commensurate gains 
in our reproductions R-MVP and R-R3M. For example, R-MVP is at par with or only slightly less performant than \unfair{MVP (EgoSoup)} on grasp affordance prediction and single-task control. We offer further discussion in \autoref{appx-subsec:preprocessing-discussion}.

\bigskip
\noindent \textbf{Q2.} \textit{Why don't you evaluate models trained with $\alpha = 1$ (pure language generation)?}

\medskip

In preliminary experiments, we partially pretrained variants of \VGen{} with values $\alpha = 0.25, 0.5, 0.75$; we focused on evaluating the downstream performance of these representations in the context of the single task visuomotor control evaluation. With $\alpha = 0.75$ we observed significant performance degradation on control tasks; furthermore, looking at the pretraining loss curves, we saw the reconstruction error plateau early in training. We found that $\alpha = 0.5$ balanced learning, and allowed us to continue to push reconstruction error down while also pushing the language generator loss (cross-entropy) lower; with $\alpha = 0.25$, we saw the opposite trend as with $\alpha = 0.75$.

\medskip

These results are to be taken with a grain of salt, given the limited pretraining duration. However, we worry that with $\alpha = 1$, we might suffer doubly for 1) never conditioning on language, which is so clearly helpful from our results, and 2) potentially fall into the same failure mode as the R-M3AE multimodal masked autoencoder from Section \autoref{sec:ablations-extensions} in the main text, overfitting to the language loss. In general, \VGen{} with $\alpha = 0.5$ already converges to a substantially higher reconstruction loss as \VCond{} and \VDual{}, as shown in the pretraining curves in \autoref{appx-subsec:pretraining-curves}. That being said, it is a promising avenue for future work to understand if this is inherent or a problem with the specific optimization procedure we used -- perhaps changing the relative scaling of the two losses over the course of pretraining may mitigate this issue, or even adaptively clipping the gradient updates depending on the relative contribution of the visual reconstructor or language generator.

\bigskip
\noindent \textbf{Q3.} \textit{Why does language during pretraining help for downstream tasks that don't use language?}

\medskip

Consider a masked visual input of a ``black, curved object above a wooden surface.'' Given this information -- and this information alone -- what is a plausible reconstruction? There are myriad objects that fit those percepts -- a black, curved object: we could lbe looking at the side of a bowl, the handle of a briefcase, the arm of a chair or stool, or in general, any number of possible options. A masked autoencoder optimizing for reconstruction must embed in the representation of this input as many of the features possible to enable good downstream reconstruction loss. It needs to model \textit{everything}, as the visual context is under-specified and ambiguous. This compressed bottleneck is core to learning a masked autoencoder, but the unfortunate byproduct of this -- in light of a vast world of possibilities -- are representations that try to capture \textit{everything} they possibly can.

\medskip

Contrast this with a world in which you are told that the same visual context is associated with the language caption  ``lifting a black coffee mug on the table.'' What changes? The posterior over possible objects collapses down to the narrow slice of possibilities captured by ``black coffee mug''; under this new set of possibilities, what does the encoder focus on? What \textit{type} of black coffee mug is on the table? If it is being lifted, \textit{how} is it being lifted? From what part of the object -- the handle (seen in frame), or somewhere else? What are the features that help further reconstruct the black coffee mug? The other nearby surfaces -- what is the mug resting on (a wooden table? the wooden arm of a chair?), is it at an angle? The additional visual context -- what type of scene are we in -- a living room, a coffeehouse? What else can I specifically encode that helps me reconstruct this cup in high-fidelity? The edges of the cup, its texture, the way the light is reflecting off of it in this particular visible context?

\medskip

Conditioning on a language description both \textit{simplifies} and \textit{focuses} what I need to represent. My encoded features are no longer general enough to cover the full range of objects that could follow from the visible context alone; instead, I can use that same capacity to represent \textit{this specific context, as denoted by language}. The encoder can focus on all of things left \textit{unspecified} by language -- arguably, the very things we \textit{want} a visual encoder for robotics to represent. Because we know that it is a ``black coffee mug,'' we can encode features around different types of black coffee mugs as a first level, and at a second level, go deeper, and actually model the \textit{low-level} features that are not  tied to semantics, but tied to core, perceptual primitives: the texture of the mug, the edges/boundaries of the object relative to other objects, even the way light reflects off of the surface. These are the features that help in tasks like grasp affordance prediction (the edges of objects), and when we learn joint representations of language and vision, the features that help with localization (grounding referring expressions) and detection. Though speculative, we can attempt to make this concrete with results: if language is indeed reducing the space over plausible reconstructions (and \textit{focusing} the encoder), we might expect lower reconstruction error when language-conditioning vs. when we condition solely on the visual context alone. This is exactly what we show in \autoref{appx-subsec:impact-of-language}, and a hint at why \Voltron{} is able to perform so strongly downstream (even without language input). The simple presence of language during pretraining refocuses the features in our representations.

\bigskip
\noindent \textbf{Q4.} \textit{Why only evaluate frozen representations? Why not fully finetune the backbones for each downstream evaluation?}

\medskip

Both MVP and R3M \citep{radosavovic2022mvp, nair2022r3m} only evaluate frozen visual representations, following a precedent set by a long tradition of work in self-supervised learning from the computer vision community \citep{chen2020simclr, radford2021clip, he2022mae}. There are two reasons for the validity of evaluating frozen representations. First, the hope is that evaluating frozen representations (via adapted per-evaluation ``heads'' on top) help us isolate the relative impact of what the representations contain -- otherwise, the separation between the standalone representations and the downstream evaluation parameters (and the co-mingling of the two when optimizing all weights via gradient descent) becomes much less clear. Second, for many of the evaluations we look at, we have extremely small amounts of data -- on the order of 1000 examples for grasp affordance prediction, 10 - 20 demonstrations for single task and language-conditioned imitation. There is a valid fear that full-finetuning the sizeable visual encoders vs. just the adaptation parameters (< 50K parameters) could lead to extreme overfitting. In general, finetuning large-scale Transformers from minimal data is an active area of research in and of itself, with work like adapters, low-rank approximations, and partial finetuning \citep{houlsby2019parameter, hu2021lora, benzaken2022bitfit}. 

\bigskip
\noindent \textbf{Q5.} \textit{Assuming pretraining datasets of (video, language) pairs feels restrictive; is there a way to leverage other sources of data?}

\smallskip

While \Voltron{} expects a dataset of videos and associated language narrations, there is a wealth of \textit{visually diverse and relevant data} that does not subscribe to this type signature:: datasets of standalone images from curated datasets \citep{deng2009imagenet, geiger2012kitti, yu2015lsun}, curated images paired with language captions as in Conceptual Captions \citep{lin2014microsoft, sharma2018conceptual}, and large in-the-wild datasets of images paired with text scraped from the internet \citep{schuhmann2021laion400m, srinivasan2021wit, schuhmann2022laion5b}. 

Luckily (though beyond the scope of this initial work), incorporating this data into the existing \Voltron{} learning pipeline is straightforward; for image data without language, we can simply ``annotate'' each example with an empty \texttt{<NULL>} token in the worst case, or alternatively, with some minimal textual metadata (e.g., a class label, dataset descriptor, or even a URL if available). To accommodate for training on variable length image contexts, a naive solution would be adopting frame dropout or padding; there are myriad ways to do this efficiently -- from Perceiver-based resampling of large patch sequences \citep{jaegle2021perceiver, alayrac2022flamingo} to different position encoding schemes \citep{su2021roformer, press2022alibi}, to more efficient attention variants \citep{beltagy2020longformer}.

\newpage

\section{\Voltron{} Implementation \& Artifacts}
\label{appx:voltron-implementation}

\bigskip
\noindent We provide complete implementation details for the various \Voltron{} models, from the small modifications to the Transformer block for added pretraining stability, to the added structural components for embedding multimodal (vision and language) inputs. All of these details are made explicit in our \href{https://github.com/siddk/voltron-robotics}{code release}, linked on our \href{https://sites.google.com/view/voltron-robotics}{project page}.

\begin{figure*}[h]
    \centering
    \includegraphics[width=\linewidth]{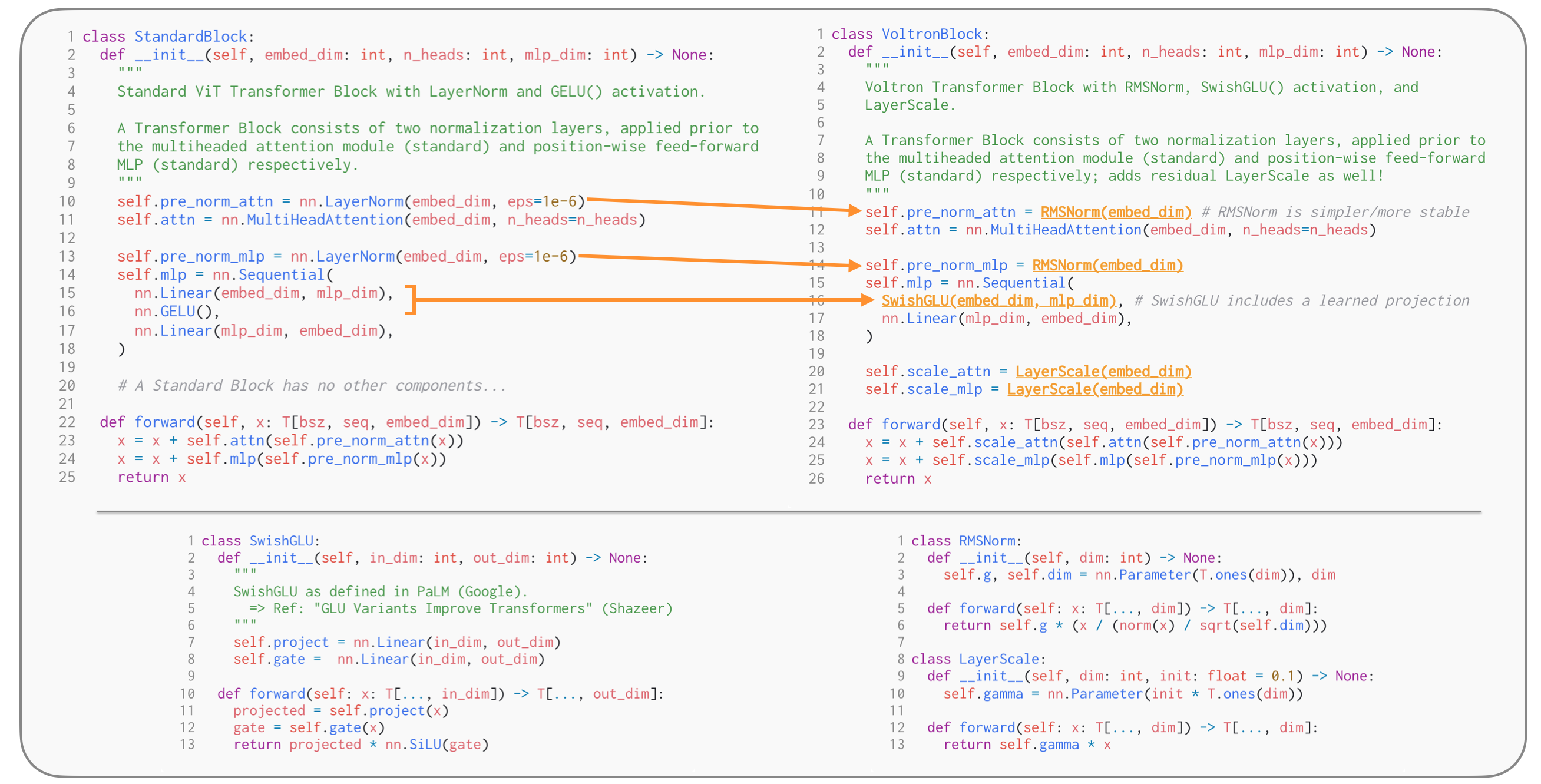}
    \vspace*{-5mm}
    \caption{\textbf{Standard vs. \Voltron{} Transformer Implementation.} The \Voltron{} Transformer Block is near-identical to the ``standard'' Transformer block used in prior work in Vision Transformers, with exceptions marked in \textcolor{xOrange}{orange}. Notably, we switch \texttt{LayerNorm} for \texttt{RMSNorm}, a standard MLP with a \texttt{GELU} activation \citep{hendrycks2016gelu} with a \texttt{SwishGLU} activation, and adopt \texttt{LayerScale} for each residual connection; these components are defined explicitly below the block definitions. In ablating these architecture modifications, we find \textit{no impact on downstream performance, but increased pretraining stability}.}
    \label{appx-fig:transformer-side-by-side}
\end{figure*}

\subsection{\Voltron{} Transformer Implementation}
\label{appx-subsec:voltron-transformer}

\noindent As mentioned in \autoref{sec:implementation-reproducibility}, we perform a series of modifications to the typical Transformer block used in prior work in the Vision Transformer and Masked Autoencoding literature to help with pretraining stability; these changes are motivated by recent work from the NLP community on training stable and performant Transformer models \citep{narang2021transformermods, karamcheti2021mistral, chowdhery2022palm}. 

\medskip

\noindent We show the side-by-side comparison of the ``standard'' Transformer block implementation vs. the \Voltron{} Transformer block in \autoref{appx-fig:transformer-side-by-side}. The changes are three-fold:
\begin{itemize}
    \item Using Root Mean-Square Normalization \citep{zhang2019rms} over the default LayerNorm; not only does RMSNorm have fewer parameters, but it has been shown to increase stability and performance \citep{narang2021transformermods}.
    \item Using the SwishGLU activation \citep{shazeer2020glu, chowdhery2022palm} over the default GELU \citep{hendrycks2016gelu}.
    \item Using LayerScale \citep{touvron2021deeper} for scaling down the magnitude of residual connections; prior work has found this to have a powerful stabilizing effect during pretraining \citep{karamcheti2021mistral}.
\end{itemize}

\medskip

\noindent We also provide pseudocode for implementing the various modifications in \autoref{appx-fig:transformer-side-by-side} (bottom); these modifications are all simple and transferable across Transformer implementations. Furthermore, as part of the no-language implementation in \autoref{sec:ablations-extensions}, we ablate the effects of these modifications on performance; we find \textit{that these modifications do not change downstream performance, but significantly increase pretraining stability}, following our initial motivation.

\subsection{Jointly Processing Vision \& Language}
\label{appx-subsec:multimodal-block-details}

\medskip

\noindent To incorporate language into the typical masked autoencoding pipeline, we add a series of small structural changes to handle 1) multi-modality, 2) sharing a Transformer decoder for both visual reconstruction and language generation, and 3) handling position encoding for both visual patch embeddings and textual tokens.

\medskip

\noindent \textbf{Multimodal Encoder.} We make the following adjustments to enable a Transformer encoder to embed multiple modalities. First, we project both our learned ``patch embeddings'' (obtained as in a standard ViT, by learning a linear transformation of our flattened RGB patches of size $p \times p \times 3$) and our pretrained language embeddings to the same space $\mathbb{R}^d$, where $d$ is the Transformer dimensionality (e.g., $d = 384$ for a ViT-Small). While we learn our patch embedding end-to-end, we initialize our language embeddings from a pretrained (and frozen) DistilBERT model \citep{sanh2019distilbert}; this is following R3M \citep{nair2022r3m}. We pad each language annotation $c$ in our dataset to a maximum length $L = 20$ tokens, additionally storing a binary length mask to ensure that each Transformer block does not attend to padding.

\medskip

Once projected into the Transformer's embedding space, we add learned modality embeddings (e.g., an embedding for \texttt{<IMG>} and \texttt{<LANG>}) to each of the respective inputs; we find that this better allows the Transformer to reason over different modalities. We initialize these learnable embeddings via a truncated normal distribution, with scale $\sigma = 0.02$, following how other special embeddings are initialized in the MAE and Vision Transformer literature \citep{he2022mae}. 

\medskip

The final step is for handling multi-frame contexts; we learn a set of frame index embeddings (e.g., for \texttt{FRAME-1}, \texttt{FRAME-2}, etc.) and add these to the corresponding patch embeddings -- i.e. we add the \texttt{FRAME-i} embedding to all patch embeddings from the first frame and so on. This further allows us to distinguish individual frame patches from one another.

\medskip

At this point, we concatenate the full sequence of flattened visual patch embeddings and language token embeddings, and feed them through the stack of Transformer blocks that form the multimodal encoder. This output is fed to the decoder, in the same fashion as a traditional masked autoencoder.

\bigskip

\noindent \textbf{Shared Transformer for Reconstruction \& Generation.} As mentioned in \autoref{sec:implementation-reproducibility}, we make one crucial change to the standard Transformer decoder in a masked autoencoder to additionally allow for language generation: namely adding a \textit{prefix mask over the language inputs} \citep{raffel2019exploring}. The goal of this mask (as stated in the main text) is to prevent information leakage when decoding; this mask selectively zeroes out dependencies in the multiheaded attention during training such that when generating language given a visual context, each language embedding at a given timestep $t$ can only attend to prior generated language at timesteps $< t$, as well as the entire visual context. This masking operates in the same way as the original decoder masking described in \citet{vaswani2013decoding}; the attention scores for all ``invalid'' inputs ($> t$) are set to 0, restricting the model from incorporating future predictions as it processes the sequence.

\medskip

Apart from this, the only other change we make to the MAE decoder is learning a separate set of modality embeddings (as described in the prior section) -- i.e. embeddings for \texttt{<IMG-DECODER>} and \texttt{<LANG-DECODER>}; the reason for this is that the Decoder sees a series of \texttt{<MASK>} embeddings representing the ``unseen'' visible context to reconstruct, as well as the new language context to generate (recall that because of the $\alpha$ gating, the language generator \textit{never} sees language embeddings from the encoder). We add these to the corresponding embeddings fed to the decoder, then resume the standard MAE decoding pipeline (reconstructing visual patches), and the language generation pipeline (autoregressively generating the original annotation). 

\bigskip

\noindent \textbf{Position Encoding.} We follow standard pratice in the masked autoencoding literature (and the same practice used by MVP), as position encode each of the patch embeddings subject to a fixed (deterministic) 2D sinusoidal embedding that reflects both vertical and horizontal positioning of each patch within a grid -- this is taken \href{https://github.com/facebookresearch/mae/blob/efb2a8062c206524e35e47d04501ed4f544c0ae8/util/pos_embed.py#L20}{directly from the original MAE codebase}. To encode text, we use a similar strategy, using a 1D sinusoidal embedding added to each token embedding in a sequence.

\begin{figure*}[t]
    \centering
    \includegraphics[width=\linewidth]{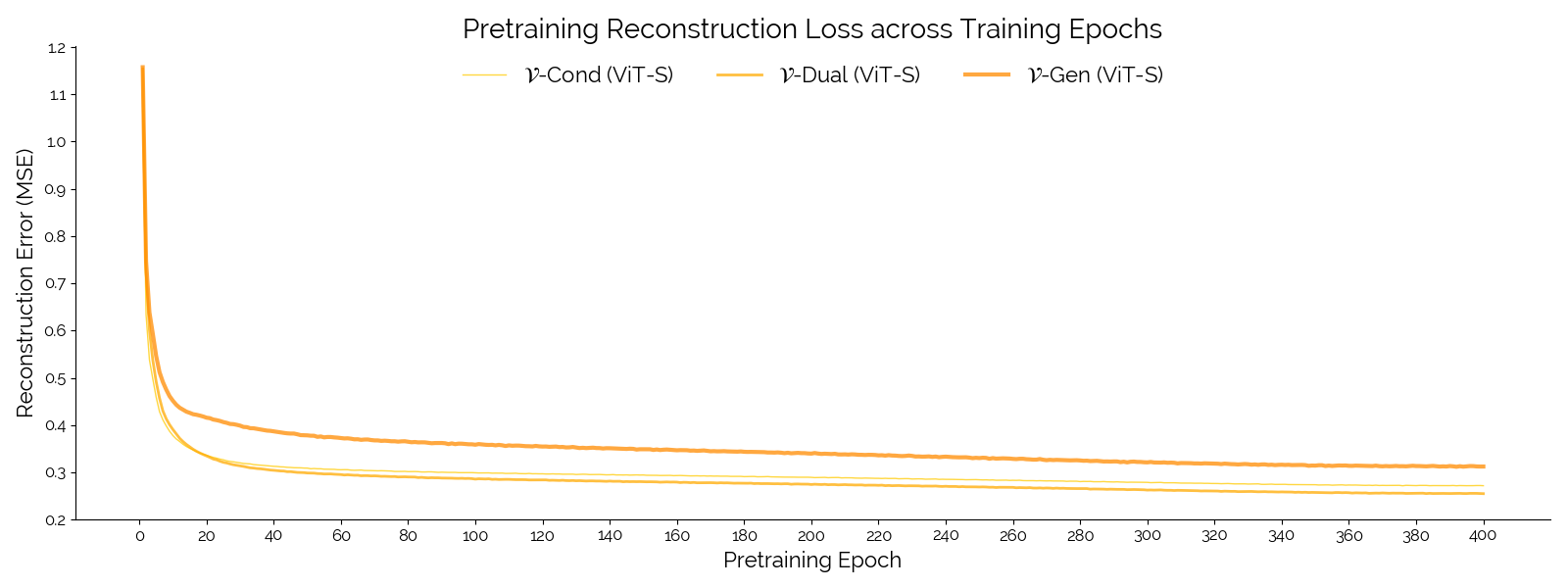}
    \vspace*{-5mm}
    \caption{\textbf{\Voltron{} Pretraining Learning Curves (Reconstruction Error).} We visualize the reconstruction error over pretraining epoch for each of the \Voltron{} models. Note that each model learns differently, converging to different reconstruction errors: both the language-conditioned models ($\alpha = 0$) converge to low reconstruction error, with \VDual{} showing that encoding and learning over multi-frame contexts allowing for a better fit. The language generative model \VGen{} ($\alpha = 0.5$) converges to a relatively higher reconstruction error, showing the tension between balancing two disparate objectives.}
    \label{appx-fig:voltron-learning-curves}
\end{figure*}

\subsection{Pretraining Curves}
\label{appx-subsec:pretraining-curves}

\noindent To further contextualize our results and enrich some of the discussion \autoref{sec:ablations-extensions} (and further on in the appendices), we include the pretraining loss curves for each of the three Voltron models we train in this work -- \VCond{}, \VDual{}, and \VGen{}. The reconstruction error curves for the three models can be found in \autoref{appx-fig:voltron-learning-curves}. In general, we find that the ``trade-off'' between language-conditioned reconstruction and visually-grounded language generation is made concrete in the pretraining loss -- both purely language-conditioned models (\VCond{}, \VDual{} with $\alpha = 0$) converge to fairly low reconstruction error; however, \VGen{} (with $\alpha = 0.5$) converges to a much higher reconstruction error -- due to the tension between optimizing for both reconstruction and language generation. We additionally note that adding even simple, dual-frame contexts enables lower reconstruction error -- even with the ViT-Small models, on the Sth-Sth dataset.

\newpage

\subsection{Index of Released Artifacts}
\label{appx-subsec:released-curves}

\noindent All of the following are linked in our \href{https://github.com/siddk/voltron-robotics}{code release} and \href{https://sites.google.com/view/voltron-robotics}{project page}:
\begin{itemize}
    \item Checkpoints for \VCond{}, \VDual{}, and \VGen{} after 400 epochs of training on Sth-Sth.
    \item Checkpoints for our reproductions R-MVP and R-R3M (both with a ViT-S and RN-50 backbone).
    \item All index files (serialized frames/order seen during training) for reproducible pretraining.
    \item Intermediate checkpoints every 20 epochs for each of the three \Voltron{} models -- along with optimizer states.
    \item Checkpoints for the ViT-Base variant of \VCond{} (86M parameters vs. 22M for a ViT-Small). 
\end{itemize}

\medskip

\noindent The \href{https://github.com/siddk/voltron-robotics}{modeling code release} additionally provides documentation and scripts for 1) training these models from scratch, and 2) downloading and extracting representations from the pretrained models. The \href{https://github.com/siddk/voltron-evaluation}{evaluation code release} provides a unified API for the various problems we evaluate on in this work.

\newpage

\section{Additional Results \& Visualizations}
\label{appx:results-visualizations}

\bigskip
\noindent We present additional results and visualizations to further support our claims from the main text. We provide additional discussion of 1) the impact of language supervision (in the context of pretraining reconstruction loss), 2) a further discussion of masked vs. generative language modeling as an objective, with an analysis of pretraining language modeling loss, 3) additional single task control results on the Adroit dexterous manipulation environments, 4) qualitative trajectory rollouts from the \VGen{} language-conditioned imitation policy, and 5) additional qualitative intent scoring results.

\begin{figure*}[h]
    \centering
    \includegraphics[width=0.95\linewidth]{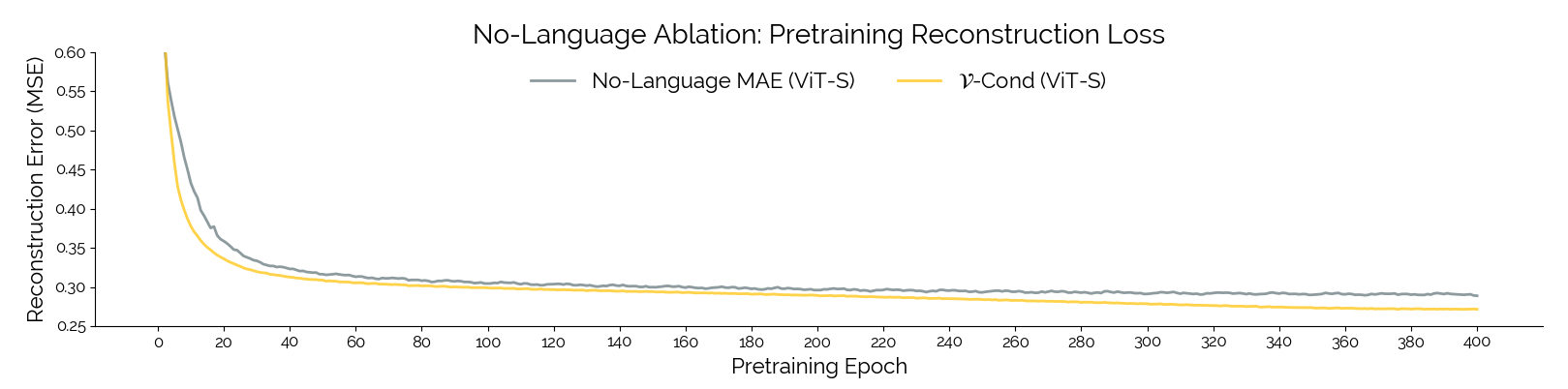}
    \vspace*{-5mm}
    \caption{\textbf{Pretraining Curves for the No-Language Ablation Experiment.} Training with language-conditioning (\VCond{}) converges to a \textit{lower reconstruction error} while also learning \textit{faster}, compared to no-language (single-frame MAE) pretraining.}
    \label{appx-fig:no-language-ablation}
\end{figure*}

\subsection{Analysis: Impact of Language-Conditioning on Reconstruction Loss}
\label{appx-subsec:impact-of-language}

\noindent As part of the ablation experiments in \autoref{sec:ablations-extensions}, we evaluate the impact of language-supervision during pretraining via a no-language ablation, training a single-frame masked autoencoder with the \Voltron{} Transformer architecture as described in \autoref{appx-subsec:voltron-transformer}; this resulting model \textit{does not condition on language at all}, but is otherwise identical to \VCond{}. In the main text, we evaluated the corresponding no-language model on a subset of evaluation tasks, showing a noticeable drop in performance across \textit{every evaluated application} (even those without language input) -- thereby showing concrete evidence as to the value of language-driven pretraining. Here we expand on those results by characterizing the behavior of both \VCond{} and the no-language ablation thereof in terms of their \textit{pretraining behavior}.

\medskip

\autoref{appx-fig:no-language-ablation} shows the reconstruction error for both \VCond{} (yellow) and the no-language ablation (gray) over the course of pretraining. There are two noticeable properties of these curves: first, \VCond{} converges to a substantially lower reconstruction error than the same model trained \textit{without language}. Second, \VCond{} is able to learn \textit{faster}, showing a steeper decline in reconstruction error earlier on in training. Taken together, these curves suggest that language-conditioning is able to focus feature learning in a way that allows the learned visual encoder to better encode masked contexts -- especially considering that the visual reconstructor is by definition \textit{not language-conditioned}. Furthermore, from the aggregate evaluation results, the features learned as a result somehow generalize better across the board, from low-level tasks like grasp affordance prediction, to high-level tasks such as control.

\begin{figure*}[h]
    \centering
    \includegraphics[width=\linewidth]{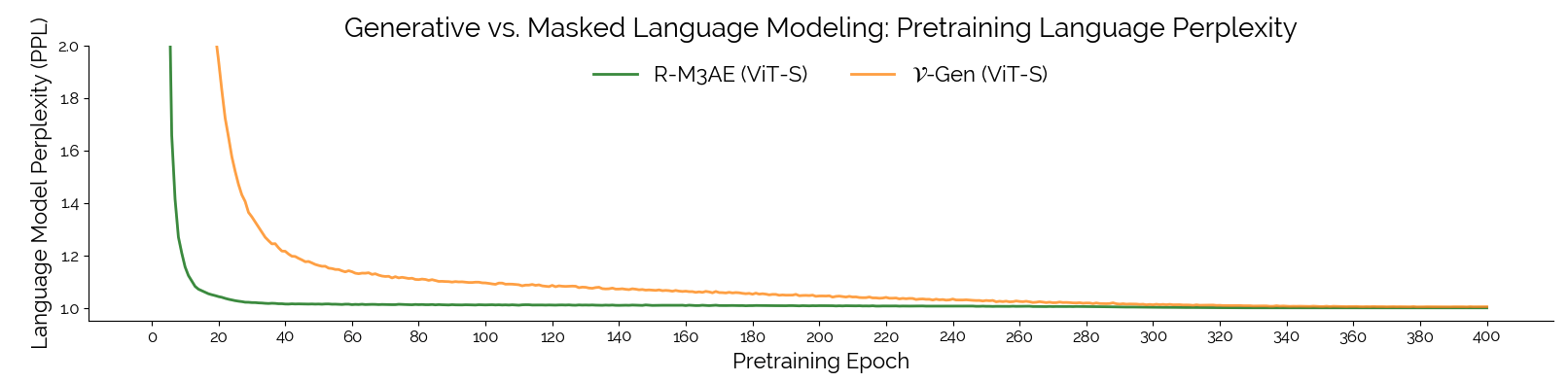}
    \vspace*{-5mm}
    \caption{\textbf{Pretraining Curves for the Generative vs. Masked Language Ablation Experiment.} Compared to multimodal masked language modeling (R-M3AE), \VGen{} ($\alpha = 0.5$) shows that with language generation as an objective, language modeling perplexity (PPL = $\exp(\text{NLL})$) gradually decreases. R-M3AE overfits to language prediction almost immediately (PPL = 1), impacting its learned representations.}
    \label{appx-fig:m3ae-pretraining}
\end{figure*}

\subsection{Analysis: Generative vs. Masked Language Modeling}
\label{appx-subsec:m3ae-results}

\noindent Later in \autoref{sec:ablations-extensions}, we raise the question: why generative (autoregressive) language modeling over masked language modeling? To help contextualize this choice, we look at recent work on combining masked autoencoders (for vision) with masked language modeling (for text), through multimodal masked autoencoders \citep[M3AE;][]{geng2022m3ae}. We reimplement this M3AE model, pretraining on the same Sth-Sth dataset used throughout this work, following the same standard of quality as for R-MVP and R-R3M. When we evaluate the corresponding R-M3AE model, we notice \textit{substantially worse performance across all evaluation domains}; in the main text we attributed this to overfitting during pretraining -- here, we provide that concrete evidence.

\medskip

\autoref{appx-fig:m3ae-pretraining} shows the \textit{language model perplexity} over time for both the R-M3AE, and the \VGen{} model (trained with $\alpha = 0.5$). Perplexity (PPL) = $\exp(\text{NLL})$ is a monotonic function of the cross-entropy loss; lower values are ``better'' with a lower bound value of 1.0. Almost immediately, the R-M3AE model overfits to the masked language modeling task, hitting a ``perfect'' perplexity of 1 (loss of 0.0) within the first 20 epochs. Contrast this with \VGen{} that learns to gradually lower perplexity of the entire course of training, almost driving down to a PPL of 1.0 by the 400th epoch. We attribute R-M3AE's poor performance to this extremely early overfitting of the language loss, again echoing the hypothesis that language generation is slightly more robust to these settings -- predict short language captions given visual context -- than a masked language modeling objective. We note that this pretraining data (Sth-Sth) is significantly different than the data used to train the original M3AE model in \citet{geng2022m3ae}; the original M3AE work used Conceptual Captions 12M \citep{sharma2018conceptual}, a rich dataset of images paired with long, descriptive captions. Further work on extending M3AE models as in \citet{liu2022instructrl} further pretrain on \textit{text-only} datasets such as Wikipedia and Toronto Books \citep{devlin2019bert} suggesting the need for diverse, broad coverage text when training (multimodal) masked language models.

\begin{figure*}[t]
    \includegraphics[width=\linewidth]{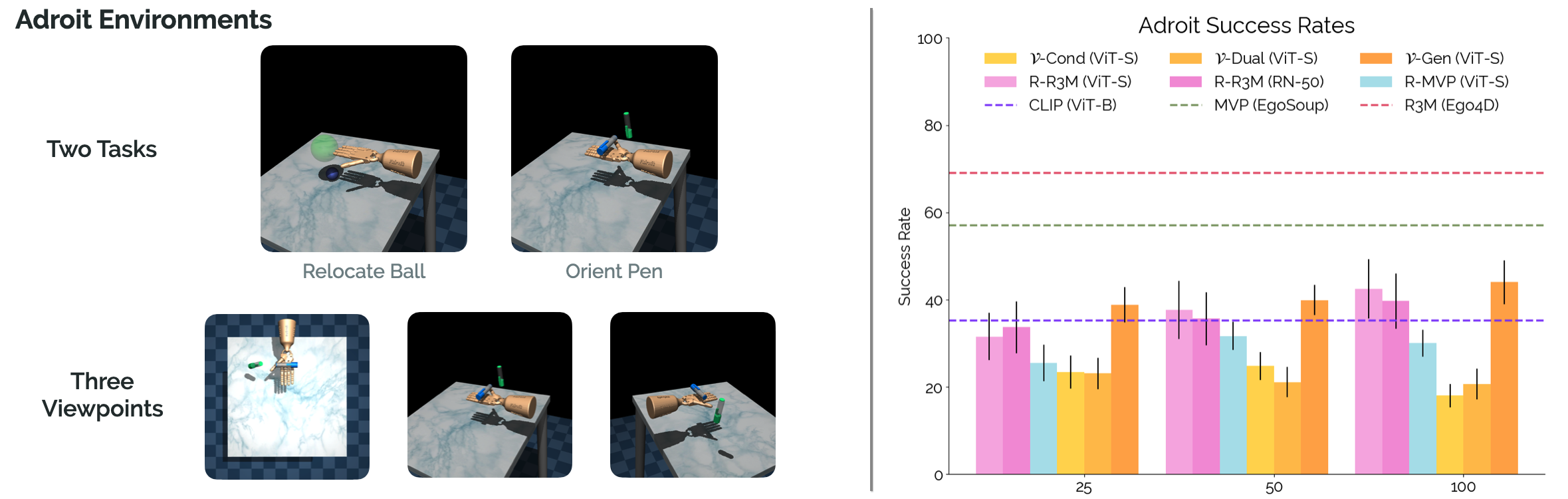}
    \vspace*{-5mm}
    \caption{\textbf{Adroit -- Single-Task Visuomotor Control Results.} Visualization of the high-dimensional Adroit environments, comprised of two dexterous manipulation tasks, with three camera viewpoints \textbf{[Left]}. Results (success rate for each of $n$ demonstrations with $n \in [25, 50, 100]$) for \Voltron{} and baselines (over 3 seeds) \textbf{[Right]}. Note the flipped trends relative to the Franka Kitchen results -- notably, the more ``high-level'' representations (from CLIP, R3M, or \VGen{}) tend to do better on this task; yet, \VGen{} is still outperforming R-R3M and CLIP, showing the benefit of language-driven flexible learning.}
    \label{appx-fig:control-adroit}
\end{figure*}

\subsection{Results: Adroit Visuomotor Control}
\label{appx-subsec:adroit-single-task-results}

\noindent To supplement our single-task visuomotor control results, we run out evaluations on the Adroit dexterous manipulation tasks from the R3M paper \citep{nair2022r3m}. The two tasks we evaluate on, depicted in \autoref{appx-fig:control-adroit} (left) consist of controlling a high degree-of-freedom robotic hand (24-DoF) for the task of 1) relocating a ball on the table to a specified target position, and 2) reorienting a pen within the hand to reach a target orientation. Given the innate difficulty of controlling a high-dimensional dexterous robotic hand over a 9-DoF fixed arm manipulator, these tasks are evaluated with $n \in [25, 50, 100]$ demonstrations instead of $n \in [5, 10, 25]$ as with the Franka Kitchen evaluation. In general, learning policies in this environment is \textit{difficult}, especially from limited data.

\medskip

Looking to the results we see that on this environment, \VGen{} and R-R3M models tend to be the most performant, in contrast with the Franka Kitchen results which favored \VCond{} and \VDual{} (the reconstruction-leaning models). Interestingly, this flipped trend seems to suggest that even within single-task control, different tasks and environments seems to prefer different visual features to perform well -- in this case, the more high-level features under models such as R-R3M and \VGen{} seem to be preferred. In a way, this makes sense; unlike with Franka Kitchen, the actual background objects and interactions thereof -- turning knobs, opening microwaves, or sliding doors with clearly marked handles -- seem more sensitive to low-level features (where on the microwave is the handle, which knob of the various possible needs to be turned). In Adroit however, these tasks are on clean backgrounds, with individual objects; the high-level behaviors instead that are more important (e.g., ``is the ball getting closer to the target location?''). It would be an interesting direction for future work to further profile other ``common'' visuomotor control tasks along this axis, to get a better understanding of what visual representations \textit{must} capture to be useful in general tasks -- to the extent of predicting ahead of time what features would be useful to aid in solving a task.

\begin{figure*}[t]
    \centering
    \includegraphics[width=\linewidth]{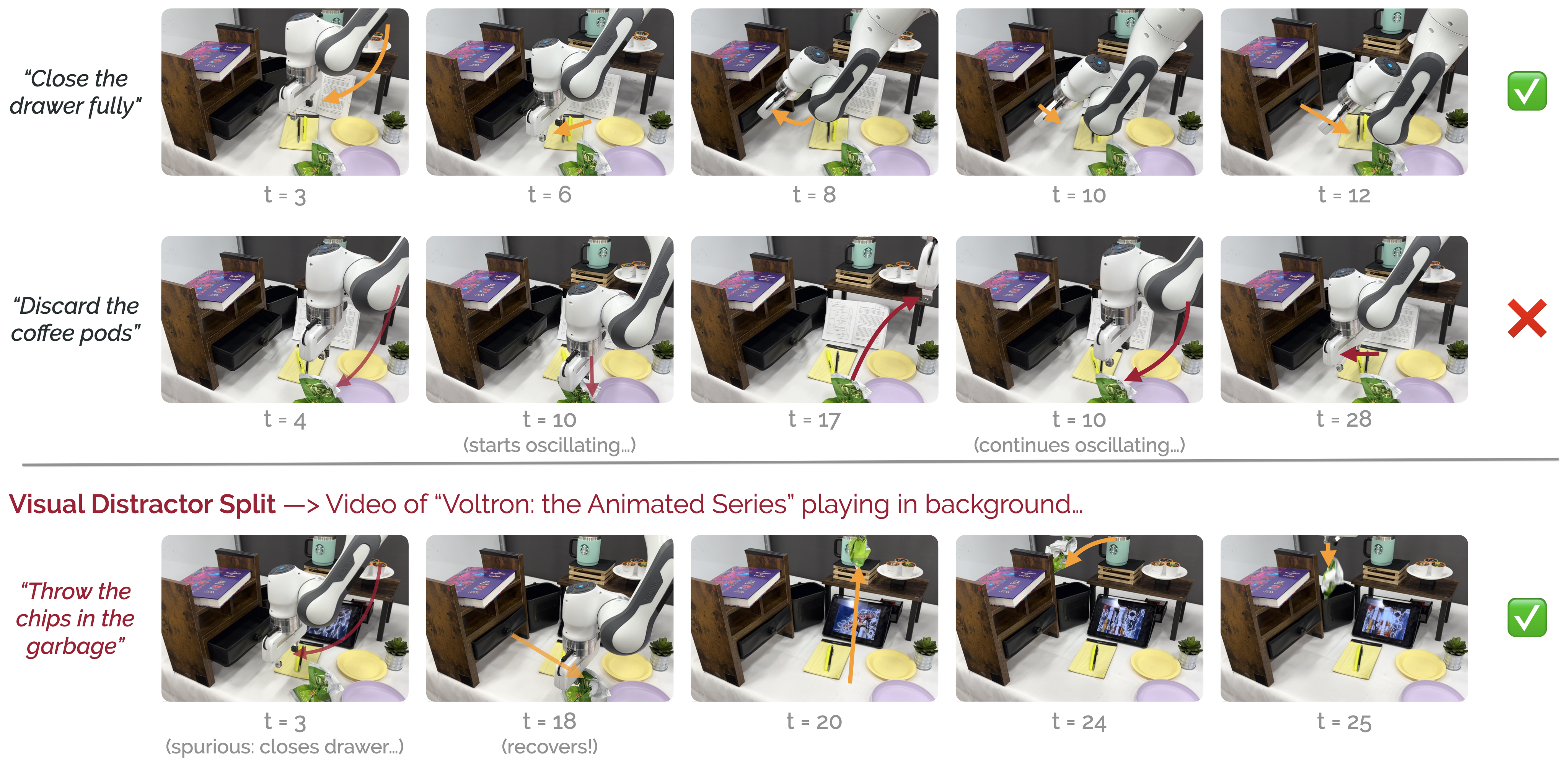}
    \vspace*{-5mm}
    \caption{\textbf{Real-World Language-Conditioned Imitation Rollouts from \VGen{}.} We visualize some rollouts from the best-performing real-world language-conditioned imitation learning model, \VGen{}. While some tasks -- e.g., discarding the plate of used coffee pods in the trash -- prove hard for all methods, \VGen{} shows smooth motion on a series of tasks, even when challenging visual distractors are present. Videos with evaluation rollouts for each method are on our \href{https://sites.google.com/view/voltron-robotics}{project page}.}
    \label{appx-fig:voltron-robot-rollouts}
\end{figure*}

\subsection{Qualitative: Real-Robot Language-Conditioned Policy Rollouts}
\label{appx-subsec:real-robot-rollouts}

\noindent While the experimental results in \autoref{sec:evaluation-suite-results} capture the quantitative success rates of various methods for language-conditioned imitation, they do not paint a picture of \textit{how} these policies behave. In \autoref{appx-fig:voltron-robot-rollouts} we show three different rollouts for the best-performing \VGen{} model: a task success (in-distribution), a task failure (in-distribution), and an example rollout from the visual distractor split. With the waypoint-based action space described in \autoref{sec:evaluation-suite-results}, we generally see smooth motions; however, the failure mode of these policies are ``oscillations'' (\autoref{appx-fig:voltron-robot-rollouts}; middle) where the policy collapses to predicting the same two waypoints repeatedly. We supplement these visualizations with \textit{full videos of rollouts from each representation learning approach} -- these are all on our \href{https://sites.google.com/view/voltron-robotics}{project page}.

\subsection{Qualitative: Additional Intent Scoring Visualizations}
\label{appx-subsec:additional-intent-scoring}

\begin{figure*}[h]
    \centering
    \includegraphics[width=\linewidth]{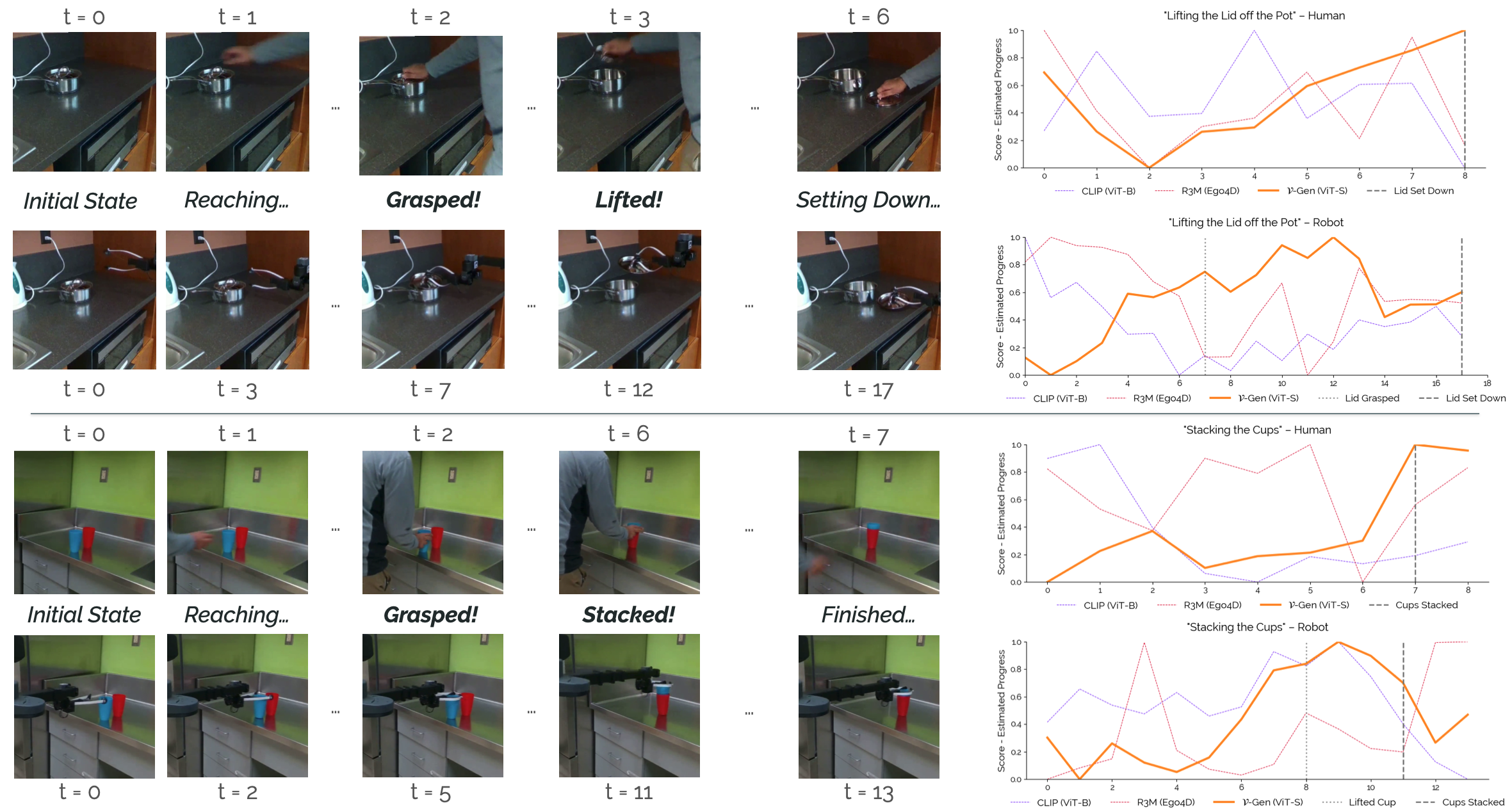}
    \vspace*{-5mm}
    \caption{\textbf{Additional Qualitative Zero-Shot Intent Scoring Examples.} Given more videos of humans and robots performing similar behaviors from the WHiRL dataset \citep{bahl2022whirl}, we evaluate the zero-shot intent scoring capabilities of \VGen{}, \unfair{R3M (Ego4D)} and \unfair{CLIP (ViT-Base)}. In general, \VGen{} continues to show
    a nuanced understanding of semantics over time, in general tracking key points in each video smoothly, whereas both baselines are for the most part predicting random scores.}
    \label{appx-fig:intent-scoring}
\end{figure*}

\noindent \autoref{appx-fig:intent-scoring} presents additional intent scoring qualitative visualizations for two other tasks from the WHiRL dataset \citep{bahl2022whirl} -- specifically ``lifting the lid off a pot'' and ``stacking cups.'' In both scenarios, we see similar behavior to the results from \S{}V of the main text: \VGen{} shows a propensity for not only tracking the key progress points in the videos for \textit{both human and robot} agents, but also providing a dense and smooth measure of intermediate progress. Both \unfair{CLIP (ViT-Base)} and \unfair{R3M (Ego4D)} unfortunately predict high-variance scores, seemingly random across the video.

\newpage

\section{Data-Equivalent Reproductions \& Reproducibility}
\label{appx:full-implementation-reproducibility}

\bigskip
\noindent In this section we provide additional discussion around two aspects of the reproduction and pretraining procedure discussed in \autoref{sec:implementation-reproducibility}: 1) preprocessing, and specifically the \textit{importance of selecting multiple images from the same context}, and 2) how to operationalize the representations from the visual encoder for downstream learning.

\subsection{Additional Preprocessing Discussion}
\label{appx-subsec:preprocessing-discussion}

\noindent We described our preprocessing approach in \autoref{sec:implementation-reproducibility}: following the R3M paper, we sample \textit{five frames from each video clip} for each epoch of pretraining. Seeing multiple frames from the same visual context is minimally necessary for the R3M time-contrastive learning objective, but we posit in this discussion (following the questions in \autoref{appx:motivating-questions}) that repeatedly sampling from the same visual context -- even with a reconstruction objective -- allows for picking up on finer-grained changes \textit{within} a context. The best evidence we have for this is in looking at how prior work constructs their pretraining datasets.

\medskip

The original MVP work \citep{xiao2022mvp, radosavovic2022mvp} constructs \textit{static datasets of images} by iterating through the various video clips in their pretraining datasets -- Sth-Sth, Ego4D \citep{grauman2022ego4d}, 100 Days of Hands \citep{shan2020hands} -- at a fixed rate, usually from 0.2 to 1 frames per second. Given video clip lengths of 2 seconds, this means that \textit{in aggregate} these pretraining datasets comprise maybe 2-3 frames sampled from the same clip, if that. Contrast that with this work and R3M, sampling multiple frames from \textit{each video clip} for \textit{every pretraining epoch} (for 400 epochs). This not only means that we are seeing the same context repeatedly, but also that we are seeing different \textit{views} of the same context; this can help tune reconstruction towards picking up on finer-grained features (e.g., if a high-capacity model is able to memorize prior contexts given enough repetition). 

\medskip

This offers a (again, speculative) explanation of why \Voltron{} models outperform \unfair{MVP (EgoSoup)} models that are both higher-capacity and trained on orders of magnitude more data -- but definitely requires further experiments to prove. In the meantime, it seems as though taking steps to use as much of the pretraining datasets we have access to as possible is in our best interest.

\newpage

\begin{figure*}[t]
    \centering
    \includegraphics[width=0.9\linewidth]{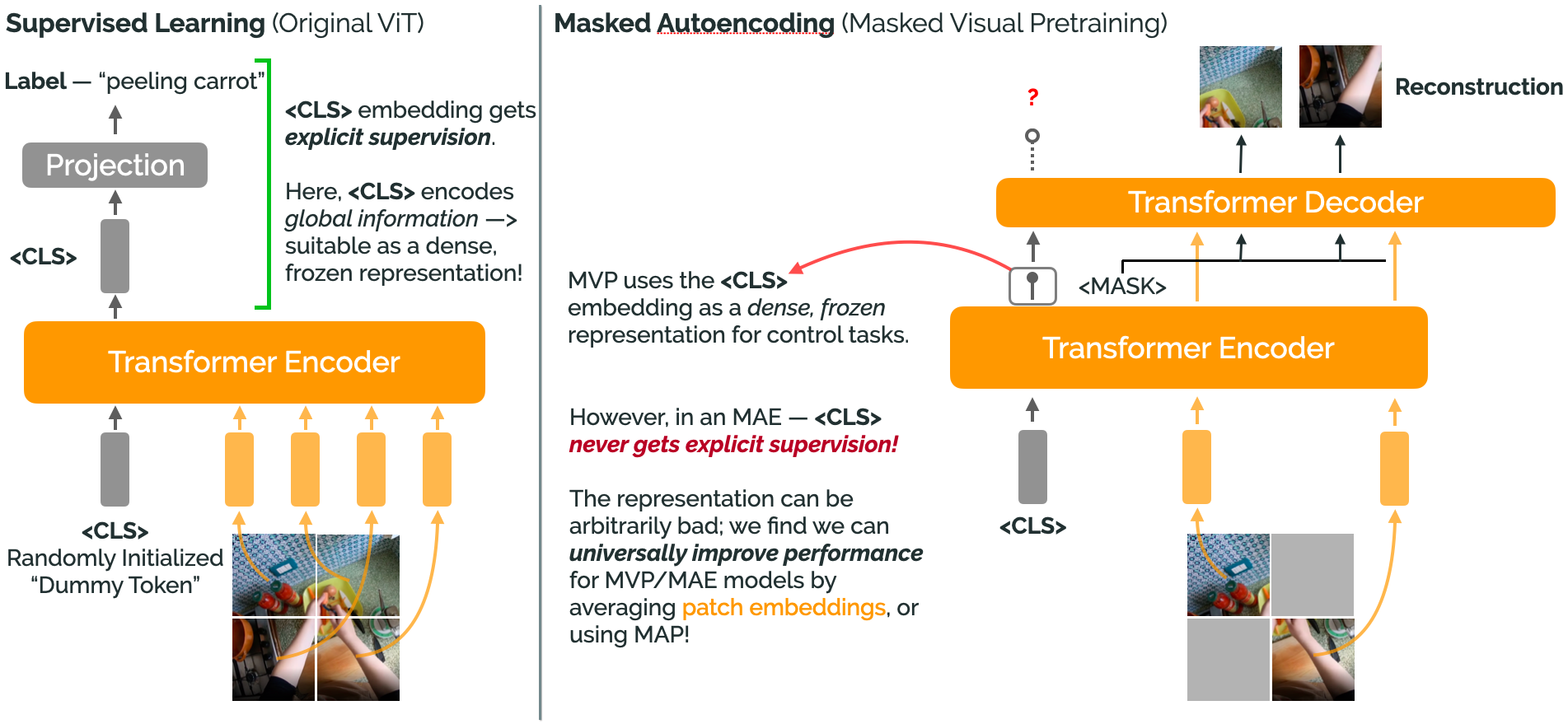}
    \vspace*{-5mm}
    \caption{\textbf{Default Feature Extraction in MAE Models.} Prior work in masked autoencoding including MVP use the embedding corresponding to a dummy \texttt{<CLS>} token appended to the Transformer input for downstream adaptation. While this is motivated in the \textit{supervised learning} setting, it is not clear what this embedding captures in the MAE setting, as it never receives explicit supervision. We find that pooling the \textit{learned patch embeddings} is strictly better.}
    \label{appx-fig:feature-extraction}
\end{figure*}

\begin{table*}[!b]
    \caption{\textbf{Feature Extraction Results.} We evaluate various feature extraction strategies on the Franka Kitchen visuomotor control tasks at $n = 10$ demonstrations. We find that multiheaded attention pooling is strictly superior for all Vision Transformer backbones; even mean-pooling over patch embeddings outperforms the default strategy from the MVP work that uses the frozen \texttt{<CLS>} embedding.}
    \label{appx-tab:feature-extraction}
    \vspace*{-2mm}
    \small
    \centering
    \begin{tabular*}{\linewidth}{@{\extracolsep{\fill}}lcccc@{}}
            \toprule
                                      & \textbf{Architecture} & \textbf{Default Extractor}       & \textbf{Mean-Pooling} & \textbf{Multiheaded Attention Pooling (MAP)}   \\ \midrule
            R-R3M                     & ViT-S         & 16.07 (Default = Mean-Pooling)           & --                    & 14.73 \\
            R-MVP                     & ViT-S         & 7.90  (Default = \texttt{<CLS>} Token)   & 9.50                  & \textbf{26.73} \\
            \VCond{}                  & ViT-S         & --                                       & 19.07                 & \textbf{27.33} \\
            \VDual{}                  & ViT-S         & --                                       & 17.40                 & \textbf{33.07} \\
            \VGen{}                   & ViT-S         & --                                       & 15.67                 & \textbf{30.33} \\ \midrule
            \VCond{}                  & ViT-B         & --                                       & 19.40                 & \textbf{30.80}  \\
            \VDual{}                  & ViT-B         & --                                       & 16.40                 & \textbf{37.27} \\
            \VGen{}                   & ViT-B         & --                                       & 15.73                 & \textbf{32.13} \\ \midrule
            \unfair{CLIP}             & ViT-B         & 17.73 (Default = Pool \& Normalize)      & 16.33                 & \textbf{22.20}  \\
            \unfair{MVP (EgoSoup)}    & ViT-B         & 18.20 (Default = \texttt{<CLS>})         & 20.13                 & \textbf{33.87} \\ \bottomrule
        \end{tabular*}
\end{table*}

\subsection{Multiheaded Attention Pooling -- Extracting Representations}
\label{appx-subsec:map-feature-extraction}

\noindent There is a critical difference between pretraining visual representations and identifying the ``right'' way to use these representations for downstream adaptation tasks. Especially for Vision Transformers trained as part of a masked autoencoder -- as mentioned at the end of Section \autoref{sec:implementation-reproducibility} of the main text -- identifying a method for extracting information from the learned representations is an open problem. The main text states -- by fiat -- that we use multiheaded attention pooling \citep[MAP;][]{lee2018settransformer} as suggested by \citet{zhai2022vitscaling} to operationalize our learned representations for our downstream tasks. Here, we further contextualize that decision with a description of alternative approaches, as well as comparative results (\autoref{appx-tab:feature-extraction}) that show the superiority of MAP-based ``feature extraction'' (referring to the process of taking the output of a Vision Transformer and producing a dense, summary vector for downstream learning) over alternative approaches.

\medskip

MVP and prior work in masked autoencoding with Vision Transformers \citep{he2022mae} make an interesting choice when it comes to extracting features: during pretraining, these works append a dummy \texttt{<CLS>} token to the input of the encoder and decoder in the masked autoencoding pipeline (depicted in \autoref{appx-fig:feature-extraction}). This ``free'' embedding is motivated by how Vision Transformers for supervised learning (e.g., classification) are parameterized: in these settings, after encoding an input image, the \texttt{<CLS>} embedding is used as (the sole) input to a linear projection into label space, thus obtaining supervision from the global loss function (e.g., the cross-entropy loss for classification). Crucially, the \texttt{<CLS>} embedding in these cases gets \textit{direct supervision} during training. However, in the masked autoencoding setting, this \texttt{<CLS>} embedding is just passed through the various Transformer layers of the encoder and decoder, \textit{never obtaining any direct or indirect supervision}; while it does attend to all other patch embeddings as a byproduct of the multiheaded attention mechanism, there is no guarantee that this embedding captures or summarize all the useful information necessary.

\medskip

Instead, recent work from the same authors of the original Vision Transformer \citep{zhai2022vitscaling} eschew the \texttt{<CLS>} embedding completely during training, instead identifying that two other strategies -- mean-pooling \textit{all} the patch embeddings output by the encoder, or using multiheaded attention pooling \citep{lee2018settransformer} -- are almost always preferable. As an aside -- this work is what motivates \Voltron{} models to also do away with the \texttt{<CLS>} embedding. 

\medskip

Multiheaded attention pooling (MAP) can be thought of as a form of cross-attention with a \textit{learned} query. Starting with a randomly initialized query vector (or optionally, \textit{set} of query vectors), a MAP block implements a shallow multiheaded attention operation, using the initialized query vector to cross-attend over the patch embeddings output by the Vision Transformer -- the resulting output is a ``weighted'' combination of the individual patch embeddings that is shaped on a per-adaptation basis. We evaluate MAP-based extraction against mean-pooling and any other ``default'' strategy (e.g., the \texttt{<CLS>} embedding used in MVP, the learned dense representation under \unfair{CLIP}) in \autoref{appx-tab:feature-extraction}. We find that MAP universally outperforms all other strategies on the Franka Kitchen control tasks (with $n = 10$ demonstrations), informing our usage of MAP as the sole feature extraction approach throughout this work. Notably, we find that MAP-based extraction when applied to the original model \unfair{MVP (EgoSoup)} released in the original work \textit{almost doubles success rate} on downstream control tasks. We even find that simple mean-pooling over patches outperforms the \texttt{<CLS>} embedding, further motivating alternate strategies.

\newpage

\section{Adapting Representations for Evaluation}
\label{appx:adaptation-for-evaluation}

\bigskip
\noindent The description of the adaptation pipeline described in \autoref{sec:evaluation-suite-results} outlines all major details for the adaptation experiments for each evaluation domain; the role of this section is to clarify any potentially ambiguous details, and further motivate some of the choices we make in implementing each evaluation. In general, all of the details for adapting representations for each evaluation in the same manner used in this work are in the released \href{https://github.com/siddk/voltron-evaluation}{evaluation code repository} that provides a unified harness for evaluating arbitrary visual representations on \textit{all evaluation domains used in this work} -- this codebase is also linked from our \href{https://sites.google.com/view/voltron-robotics}{project page}.

\medskip

In general, for each evaluation domain, we keep the adaptation architecture as simple as possible, and optimization parameters simple as well. For all applications we use an AdamW optimizer \citep{kingma2015adam} with the default learning rate of 1e-3, and weight decay of 0.01.

\bigskip

\noindent \textbf{Grasp Affordance Prediction.} We implement the adaptation head for the grasp affordance prediction task following recent work in learning segmentation heads on top of vision transformer features, specifically following the procedure outlined in Segmentation Transformers via Progressive Upsampling (SETR-PUP) \citep{zheng2021setr}. A PUP block is straightforward -- we first extract all patch embeddings from the output of our Vision Transformer encoder, using a shallow MAP block with the same number of seed vectors as patches output by the encoder. We then reshape the extracted features into a \textit{grid}, then stack a series of 4 upsampling blocks (channel depths of $[128, 64, 32, 16]$, ReLU activation) that consist of a 2D convolution followed by a bilinear upsampling, until we recover a grid of the same size of the original image. We finally apply a spatial softmax, predicting distributions over each of the possible labels (``graspable,'' ``non-graspable,'' ``background''), and compute our loss per-pixel. We optimize with a batch size of 64, for 50 epochs in total. Given the small size of the dataset, we find that there is a great deal of variance across random initializations; we report results by running 5-fold cross-validation, taking the model with the best performance across validation folds to compute final test statistics.

\bigskip

\noindent \textbf{Referring Expression Grounding.} We use a simple adaptation head for referring expression grounding that extracts a single dense representation from our learned encoder via a shallow MAP block with a single seed vector (the default extractor for obtaining a vector representation of a visual input). For representations that are not language-conditioned, we concatenate this vector with the language embedding under the appropriate model -- e.g., the CLIP text embedding for \unfair{CLIP (ViT-Base)} -- or the DistilBERT language embedding for pure visual models (e.g., MVP). We then feed this context through a 4-layer MLP (hidden dimensions of $[512, 128, 128, 64]$, GELU activation) that directly predicts bounding box coordinates as $(x, y, \text{width}, \text{height}$). We use a Huber loss to compute error. We optimize with a batch size of 512, for 10 epochs in total, using the provided validation set for model selection.

\bigskip

\noindent \textbf{Single-Task Visuomotor Control.} We first extract a dense representation using a shallow MAP block (as described above), then follow the exact procedure for evaluating both Franka Kitchen and Adroit policy learning as described in the R3M work \citep{nair2022r3m}. Namely, we concatenate the visual representation with the robot's proprioceptive state, followed by a BatchNorm layer \citep{ioffe2015batch}. These are then fed to a 2-layer MLP ($d = 256$) that directly predicts action targets for computing mean-squared error against the ground-truth actions. Following R3M, we run 20,000 gradient steps with a batch size of 32, evaluating the models online every 5000 steps on a heldout set of 50 environments (fixed seed) -- we report success rate subject to the best performing model from the online evaluation. We run three seeds for each combination of viewpoint, number of demonstrations, and task.

\bigskip

\noindent \textbf{Real-World Language-Conditioned Imitation.} The full set of language instructions generated by ChatGPT can be found on our \href{https://sites.google.com/view/voltron-robotics}{project page}. For adaptation, we first extract a representation as with the referring expression evaluation by using a shallow MAP block, and concatenating the corresponding language embedding as appropriate. We concatenate this fused vector with the robot's proprioceptive state, and pass the corresponding embedding to a BatchNorm layer. Then, following recent work on real-world imitation learning \citep{mandlekar2021robomimic}, we only train a shallow 2-layer MLP with ($d = 64$) to predict action targets for computing mean-squared error against the ground-truth waypoint actions. We optimize with a batch size of 256, and train for 10 epochs. As policy evaluation in the real-world is expensive -- especially for the five approaches we evalaute -- we uniformly choose the last epoch checkpoint to perform evaluation rollouts. 

\bigskip

\noindent \textbf{Qualitative: Zero-Shot Intent Scoring.} This is a zero-shot evaluation with no adaptation data, only applicable to the representation learning models capable of ``scoring'' joint vision-language contexts: \VGen{}, \unfair{CLIP (ViT-Base)}, and \unfair{R3M (Ego4D)}. We download videos from the WHiRL dataset off of the WHiRL website: \url{https://human2robot.github.io/}. To generate plots, we sample frames at 2 FPS from each video, center cropping and resizing each frame prior to passing it to each model.

\end{document}